%% file: main.tex
\documentclass[a4paper,fleqn]{cas-dc}

\usepackage{amssymb,mathtools}
\usepackage{algorithm}
\usepackage{algpseudocode}
\usepackage{booktabs}
\usepackage{array}
\usepackage{xcolor}
\usepackage[table]{xcolor}
\usepackage{array}
\usepackage{url}
\usepackage{makecell}
\usepackage{graphicx}
\usepackage{siunitx}
\usepackage{float}
\usepackage[authoryear]{natbib}
\usepackage{amsmath,bm}
\usepackage{subcaption}
\usepackage{xspace}
\usepackage{microtype}
\usepackage{tabularx}
\usepackage[capitalise,noabbrev,nameinlink]{cleveref}
\newcommand{\apply}{\mathbin{\odot}}

\DeclareMathOperator*{\argmin}{arg\,min}
\DeclareMathOperator{\dist}{dist}

\DeclareMathOperator{\tr}{tr}
\DeclareMathOperator{\clip}{clip}
\newcommand{\R}{\mathbb{R}}

\newcommand{\SE}{\mathrm{SE}}
\newcommand{\Sim}{\mathrm{Sim}}
\newcommand{\Tmat}{\mathbf{T}}
\newcommand{\Smat}{\mathbf{S}}
\newcommand{\Rmat}{\mathbf{R}}
\newcommand{\tvec}{\mathbf{t}}
\newcommand{\nvec}{\mathbf{n}}
\newcommand{\pvec}{\mathbf{p}}
\newcommand{\qvec}{\mathbf{q}}
\newcommand{\xvec}{\mathbf{x}}

\newcommand{\zvec}{\mathbf{z}}
\newcommand{\dvec}{\mathbf{d}}
\newcommand{\mvec}{\mathbf{m}}
\newcommand{\cvec}{\mathbf{c}}

\newcommand{\I}{\mathbf{I}}
\newcommand{\Pmat}{\mathbf{P}}
\newcommand{\rhoHuber}{\rho_{\delta}}
\newcommand{\norm}[1]{\left\lVert #1 \right\rVert}

\newcommand{\mm}{\,\mathrm{mm}}
\newcommand{\s}{\,\mathrm{s}}
\newcommand{\ZED}{ZED\xspace}

\crefname{figure}{Fig.}{Figs.}
\Crefname{figure}{Figure}{Figures}
\crefname{table}{Table}{Tables}
\Crefname{table}{Table}{Tables}
\crefname{equation}{Eq.}{Eqs.}
\Crefname{equation}{Equation}{Equations}

\usepackage{cleveref}
\usepackage{svg} 
\def\tsc#1{\csdef{#1}{\textsc{\lowercase{#1}}\xspace}}
\tsc{WGM}
\tsc{QE}
\begin{document}
\let\WriteBookmarks\relax
\def\floatpagepagefraction{1}
\def\textpagefraction{.001}
\setlength{\abovecaptionskip}{4pt}
\setlength{\belowcaptionskip}{0pt}
\setlength{\floatsep}{8pt plus 2pt minus 2pt}
\setlength{\textfloatsep}{10pt plus 2pt minus 2pt}

\shorttitle{Marker-free AR margin localization}
\shortauthors{Y. Yang et~al.}

\title[mode = title]{Marker-free deformable registration and fusion for augmented reality-guided positive margin localization during tumor resection surgery}

\author[1]{Yue Yang}
\credit{Conceptualization, Methodology, Software, Validation, Writing, Visualization, Investigation, Data curation, Formal analysis}

\author[1]{Annie Benson}
\credit{Investigation, Data curation, Writing -- review \& editing}

\author[1]{Matthieu Chabanas}
\credit{Investigation, Project administration, Methodology, Supervision, Writing -- review \& editing}

\author[2]{Jason Slagle}
\credit{Methodology, Investigation, Writing -- review \& editing}

\author[2]{Thomas Myles}
\credit{Investigation, Validation, Writing -- review \& editing}

\author[2]{Matthew B. Weinger}
\credit{Methodology, Supervision, Writing -- review \& editing}

\author[1]{Jon S. Heiselman}
\credit{Software, Writing -- review \& editing}

\author[1]{Michael I. Miga}
\credit{Supervision, Funding acquisition, Writing -- review \& editing}

\author[2]{Michael Topf}
\credit{Conceptualization, Resources, Supervision, Funding acquisition, Writing -- review \& editing}

\author[1]{Jie Ying Wu}[orcid=0000-0002-7306-8140]
\cormark[1]
\credit{Project administration, Conceptualization, Methodology, Resources, Supervision, Funding acquisition, Writing -- review \& editing}

\affiliation[1]{
    organization={Vanderbilt University},
    addressline={1400 18th Avenue South},
    city={Nashville},
    state={Tennessee},
    postcode={37212},
    country={USA}
}

\affiliation[2]{
    organization={Vanderbilt University Medical Center},
    addressline={1211 Medical Center Drive},
    city={Nashville},
    state={Tennessee},
    postcode={37232},
    country={USA}
}

\cortext[cor1]{Corresponding author (jieying.wu@vanderbilt.edu).}

\begin{abstract}
Positive margins in head and neck oncologic surgery require mapping specimen-side pathology findings to the patient resection bed. This is challenging because pathologists identify the positive margin on slices of resected and deformed specimen, while surgeons must relocate the corresponding site on the resection bed using only verbal descriptions and no visual guidance. We present a marker-free augmented reality (AR) workflow for mapping a margin label from a three-dimensional specimen scan to the resection bed. The method combines contour-constrained deformation, residual alignment to a depth scan, surface-based fusion to a head-mounted display, and target projection onto the reconstructed bed. Bead-suture correspondences estimate specimen deformation, whereas patient-to-display fusion does not require external fiducial markers. Following formative experiments, five residents and surgeons performed cadaveric cheek and scalp re-resection tasks under verbal guidance, verbal guidance with specimen examination, and AR guidance. Deformation target errors were 7.63 ± 3.74 mm for the cheek and 3.72 ± 1.02 mm for the scalp; residual specimen-to-bed distances were 2.43 ± 2.15 mm and 2.19 ± 1.06 mm, respectively. Fusion error did not differ significantly between marker-free and marker-based methods on either cadaver; overall marker-free fusion error was 2.15 ± 0.87 mm. End-to-end margin localization error decreased from 21.40 ± 3.84 mm with verbal guidance and 16.09 ± 4.30 mm with specimen examination to 6.19 ± 1.79 mm with AR guidance ($p < 0.001$). Online fusion required 5.23 ± 0.34 s. These results demonstrate effective marker-free AR guidance for positive-margin localization and support more precise tumor resection.
\end{abstract}


\maketitle

\section{Introduction}
\label{sec:introduction}

Head and neck squamous cell carcinoma (HNSCC) creates a large global cancer burden, with nearly 890,000 new cases estimated worldwide \citep{sung2021global}. Surgery remains a central treatment for many head and neck tumors, and its effectiveness depends strongly on complete tumor resection with negative margins \citep{sunkara2023association}. Negative margins are especially difficult to achieve in HNSCC where a few millimeters can separate residual tumor from nerves, vessels, facial skin, or critical cosmetic goals. Positive, involved, and close margins have been linked to higher local recurrence and worse survival in mucosal HNSCC \citep{wong2012influence,alicandri2013surgical,hamman2022impact}, which is a common and clinically important topic \citep{prasad2023trends,prasad2024reresections}. Margin status also guides postoperative treatment. Two landmark randomized trials showed the importance of high-risk pathologic features, including positive margins, for selecting postoperative chemoradiotherapy in locally advanced head and neck cancer \citep{bernier2004postoperative,cooper2004postoperative}.

Intraoperative frozen section analysis helps surgeons detect positive or close margins while the patient is still in the operating room. Specimen-oriented intraoperative margin assessment and modern frozen section studies support the value of acting on margin information during the same operation \citep{horwich2021specimen,nentwig2021impact,ali2024diagnostic}. When a pathologist identifies a positive margin on the resected specimen, the surgeon must relocate that site on the patient and remove additional tissue from the correct part of the resection bed. The head and neck region, especially inside the oral cavity, has complex three-dimensional (3D) anatomy, and the specimen is detached, rotated, inked, and sectioned, and the pathology results are usually described through a short verbal or written anatomic label. Conventional margin labeling can lead to substantial localization variability, and recent relocation studies show that this problem remains clinically meaningful \citep{kerawala2001relocating,banoub2023variance,miller2024howfar}. Thus, surgeons can miss the residual disease site even when they act on the frozen section result. In a recent study, only 29\% of re-resections after an initial positive margin contained additional carcinoma, carcinoma \textit{in situ}, or severe dysplasia \citep{prasad2024reresections}. Imprecise re-resection may leave residual tumor behind or lead to unnecessary removal of healthy tissue when the surgeon widens the resection to compensate for uncertainty.

3D specimen mapping has emerged as a practical way to preserve spatial information during pathology processing. Structured-light or photogrammetry-based scanning can create a textured model of the resected specimen, and digital annotation can record sampling locations, inked surfaces, section planes, and positive or close margin sites \citep{sharif2023cad,saturno2023scanning,miller2024virtual}. These digital maps improve communication between surgeons and pathologists, yet they must be viewed outside the operative field with the surgeon mentally transferring the information back to the field. Augmented reality (AR) offers an attractive solution by overlaying specimen-derived margin information onto the resection bed. Medical AR has a long history in image-guided surgery, and several reviews have described its potential to reduce mental mapping, improve hand-eye coordination, and return 3D information to the surgeon's line of sight \citep{sielhorst2008advanced,kersten2013state,bernhardt2017status,birlo2022utility}. Recent head and neck studies have shown that AR can guide re-resection in cadaveric or phantom settings, including systems that upload a 3D scanned specimen into a HoloLens environment \citep{prasad2023ar,tong2024development, yang2026all}. Mixed-reality and navigation systems have also supported tumor resection in oral and maxillofacial surgery \citep{tang2022accuracy}. While these studies support the concept, most current workflows still depend on manual alignment, external navigation hardware, fiducial markers, or rigid specimen assumptions.

A complete AR system for positive margin localization should solve three connected technical problems. \textit{First}, it should model the deformation of the resected tissue. Fresh specimens do not remain conformationally identical to the resection bed after excision. They shrink and deform because of release of tissue tension, electrocautery effects, gravity, handling, and fixation \citep{mistry2005shrinkage,burns2021shrinkage,kshithi2022formalin}. General deformable registration methods, such as thin-plate spline models and coherent point drift, provide useful mathematical tools for nonrigid alignment \citep{bookstein1989principal,myronenko2010point}. However, positive margin localization needs deformation models that preserve the specimen surface and resection contours, because positive margins are defined on the resection boundary. A model that fits only sparse fiducials can still place edge regions poorly \citep{heiselman2024image, richey2022computational}, which may move the target away from the true resection-bed location. The deformation step therefore should use the fiducials to anchor the specimen while also using the resection contour to control the boundary where clinical decisions are made.

\textit{Second}, the system should ideally register the deformed specimen back to the patient. Marker-based navigation can provide reliable coordinates, but it adds setup time, line-of-sight constraints, sterile-field management, and hardware burden. Manual hologram alignment avoids hardware but introduces user-dependent error and makes quantitative evaluation difficult. Surface-based point cloud registration offers a marker-free alternative. Iterative closest point and point-to-plane variants remain widely used for local refinement, while feature descriptors and robust global registration methods improve initialization when there are outliers or partial overlap \citep{Besl1992ICP,chen1992object,rusu2009fpfh,yang2021teaser}. However, surgical resection beds differ from common registration benchmarks. Drapes can introduce large flat regions, blood and moisture can affect depth measurements, and the useful geometry may concentrate along the resection contour rather than across the whole field. A clinically useful registration method should therefore down-weight uninformative planar regions and emphasize stable resection edges, curvature, and local surface changes.

\textit{Third}, the system should fuse the registered margin location into the AR world coordinate frame and quantify each error source. This need for error quantification applies to any AR guidance platform. Optical see-through head-mounted displays (HMDs) are particularly attractive because they keep the surgeon's view on the patient, but their depth sensors have limited accuracy compared with dedicated scanners and tracked tools \citep{hubner2020evaluation,kerkhof2025depth}. For margin guidance, a few millimeters of error may change whether a re-resection targets the correct edge. Therefore, the evaluation should not report only a final overlay error. It should measure deformation accuracy, specimen-to-bed registration, AR fusion, latency, end-to-end targeting accuracy, and user feedback in a surgical setting. This component-level analysis makes it possible to identify which part of the workflow limits performance and which part can be improved for clinical adoption.

In this paper, we introduce a marker-free AR guidance system that combines deformation modeling, surface-based registration, and HMD-based visualization for positive margin localization during tumor re-resection. We compared the proposed system with the current clinical approach in a simulated cadaveric surgery study with head and neck residents and surgeons targeting cutaneous tumor sites on the cheek and scalp. We evaluated the system with metrics that test each component, including deformation and registration, AR fusion, end-to-end accuracy, latency, and subjective user experience. Our main contributions are as follows:
\begin{itemize}
    \item We developed and integrated deformation modeling, marker-free registration, and AR HMD visualization for positive margin localization.
    \item We designed a clinical workflow tailored to head and neck tumor re-resection that fits within the intraoperative pathology time window and avoids external patient-to-HMD markers.
    \item We quantified the AR guidance system with component-level metrics, end-to-end targeting accuracy, latency, and surgeon user feedback.
\end{itemize}

\section{Method}
\label{sec:method}

\subsection{AR workflow development}
\label{sec:workflow}

Current intraoperative margin management and localization follow a verbal
communication workflow. The surgeon performs the primary resection and sends
the specimen, or selected margins, to pathology for frozen section or other
rapid pathologic assessment. Specimen-oriented margin assessment has gained
support because it links the pathologic result to the resected tissue, yet the result often returns to the operating room as a vague anatomic description rather than a patient-side spatial target \citep{horwich2021specimen,brandwein2023radical}. A typical message may state that a positive margin lies ``in the anterior-inferior region'' of the
specimen. The surgeon must then map that information back to the resection bed using memory, orientation sutures or inks, and discussion with the
pathologist. Recent studies show that conventional anatomic labels create
localization variance, and re-resections after positive margins often fail to contain additional tumor cells \citep{banoub2023variance,miller2024howfar,prasad2024reresections}. This gap
motivates an improved workflow that preserves the current clinical sequence
but replaces the final mental transfer with an AR-guided spatial transfer.

\begin{figure*}[t]
    \centering
    \includegraphics[width=\textwidth]{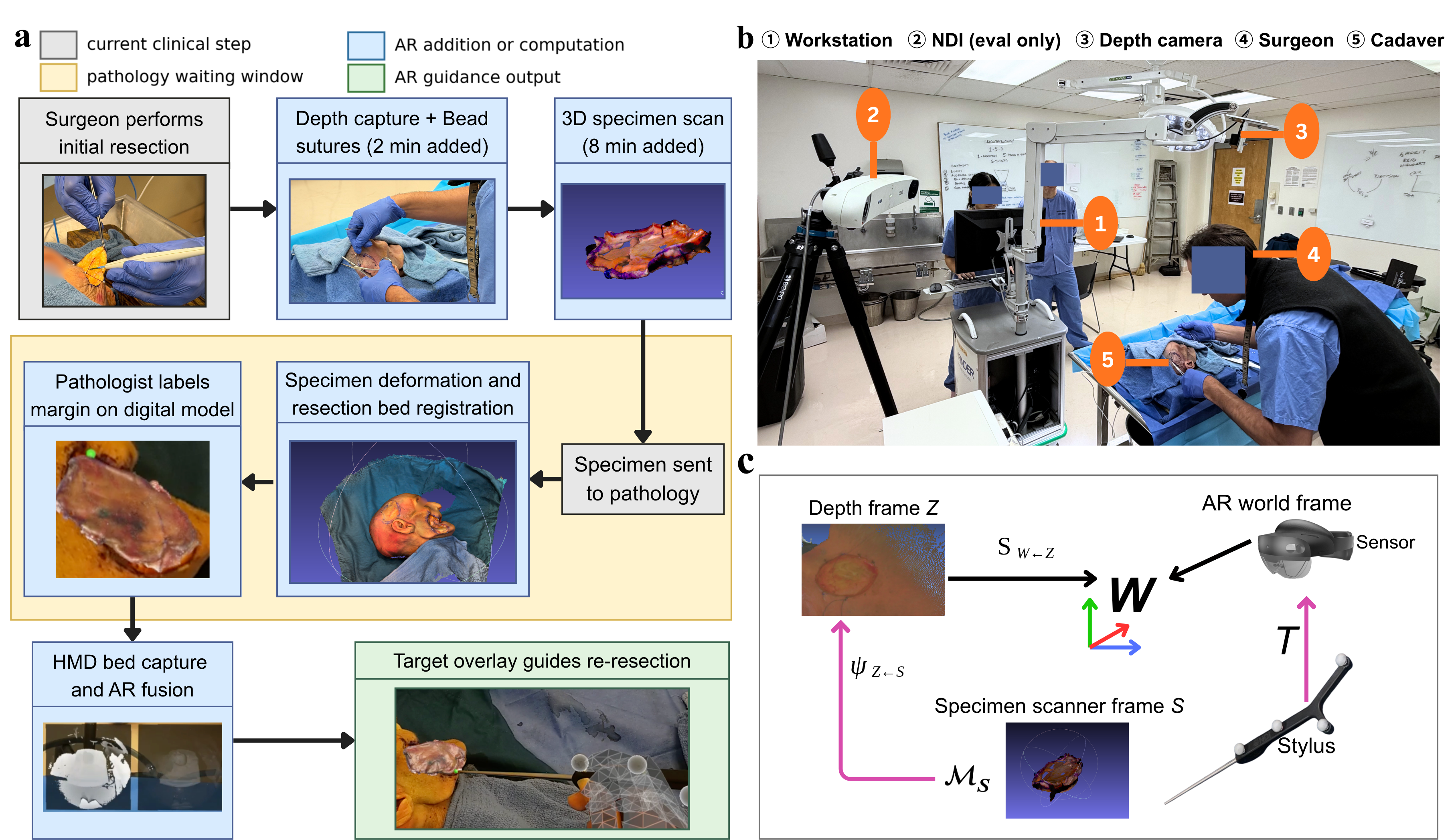}
    \caption{Workflow of the proposed system.
    \textbf{(a)} Proposed AR-guided workflow integrated with the current
    clinical workflow. Gray boxes denote existing clinical steps, blue boxes
    denote AR additions or computations, the pale yellow region denotes
    processing performed during pathology analysis, and the green box denotes
    AR guidance. The workflow adds approximately 10 min for bead-suture
    placement and 3D specimen scanning; deformation, resection-bed
    registration, and AR preparation run during pathology analysis.
    \textbf{(b)} Experimental setup; NDI tracking was used only for
    evaluation.
    \textbf{(c)} Coordinate transfer among components.}
    \label{fig:workflow}
\end{figure*}

\cref{fig:workflow} shows the current standard workflow with the proposed
AR-guided workflow. The clinical setup places a depth camera above the
surgical bed to acquire a dense point cloud of the resection bed after the
primary excision. The specific depth camera used in the final workflow was
selected in the formative experiment described in
\cref{subsec:depth_camera_selection}. Immediately after resection, the surgeon sutures four small beads along the resection line onto the resection bed and their paired sutures along the edge of the resected specimen. These bead-suture pairs ensure that fiducials appear in the 3D specimen scan and the external depth-camera acquisition, and they act as deformation fiducials that connect the scanned specimen boundary to the resection-bed contour. Because these beads are attached to a deformable resection boundary rather than a rigid, spatially distributed reference frame, they are unsuitable for stable patient-to-HMD registration, and the AR fusion step therefore remains marker-free.

After leaving the surgical field, the specimen is transferred for pathologic
evaluation. The proposed AR-guided workflow acquires a 3D specimen scan before routine gross pathologic handling. The 3D scanner produces a textured mesh that preserves the specimen surface, color, and boundary shape. Prior 3D specimen mapping work has shown that optical scanning can fit into head and neck margin communication workflows \citep{sharif2023cad,saturno2023scanning}.

The digital specimen then becomes the input to deformation modeling. The model uses bead-suture correspondences and resection-bed contour information to deform the specimen mesh toward the patient-side geometry. The deformed specimen is registered to the external depth-camera resection-bed point cloud with surface alignment, which produces a common depth coordinate frame containing the resection bed, the deformed specimen, and any margin label that pathology later adds. These computational steps can occur while pathology performs margin analysis, limiting additional delay in the operative workflow.

When pathology identifies a positive or close margin, the pathologist labels
the site directly on the digital specimen rather than communicating only an
anatomic phrase. This site can be represented as a virtual 3D target object.
The margin-labeled model is then loaded onto an AR HMD (HoloLens 2, Microsoft, USA). The surgeon wears the HMD and looks at the resection bed. The online fusion pipeline captures a local resection-bed point cloud from the HMD depth sensor and aligns it to the external depth-camera resection-bed point cloud that already carries the registered deformed specimen. This alignment maps the external depth frame into the HMD world frame. After fusion, the system renders the deformed specimen as a semi-transparent overlay and displays the positive-margin labels as green spherical targets on the resection bed. The semi-transparent specimen gives the surgeon a visual consistency check, while the green targets emphasize the locations that require re-resection.

Thus, the workflow maps a pathology label from the specimen-scanner frame
$S$ to the HMD world frame $W$ through the external depth-camera frame $Z$.
The AHAT depth-camera frame is denoted by $H$. The primary data objects are
the textured specimen mesh $\mathcal{M}_S$, the external depth resection-bed
cloud $\mathcal{B}_Z$, the HMD-captured bed cloud $\mathcal{B}_W$, and the
local HMD depth-correction reference set $\mathcal{L}_W$. The boundary and
rim sets $\Gamma_S$, $\Gamma_Z$, and $\Gamma_W$ are derived from these data.
Four bead-suture correspondences
$\mathcal{A}=\{(\mathbf{a}_k^S,\mathbf{a}_k^Z)\}_{k=1}^{4}$ are used only to
initialize and constrain specimen deformation; they are not used for
patient-to-HMD fusion. The reference points in $\mathcal{L}_W$ are used only
to estimate a local correction of HMD depth and are not used to align the
patient or specimen during online fusion.

For a rigid transform $\Tmat_{A\leftarrow B}\in\SE(3)$,
$\Tmat_{A\leftarrow B}\apply\xvec^B$ denotes left multiplication of the
homogeneous point $[\xvec^{B\top},1]^\top$, followed by removal of the final
homogeneous coordinate. The same action notation is used for a similarity
transform $\Smat_{A\leftarrow B}\in\Sim(3)$. \cref{tab:notation} lists the data and maps used repeatedly in the method.

\begin{table*}[t]
\centering
\caption{Core data and maps used in the registration and fusion pipeline.}
\label{tab:notation}
\small
\renewcommand{\arraystretch}{1.08}
\setlength{\tabcolsep}{5pt}
\begin{tabularx}{\textwidth}{@{}p{0.25\textwidth}X@{}}
\toprule
\rowcolor{gray!18}
\textbf{Symbol} & \textbf{Definition} \\
\midrule

\rowcolor{gray!7}
$\mathcal{M}_S$
& Textured specimen mesh and specimen-side labels in scanner frame $S$. \\

$\mathcal{B}_Z,\mathcal{B}_W$
& Post-resection bed point clouds in external depth frame $Z$ and HMD world
frame $W$. \\

\rowcolor{gray!7}
$\mathcal{L}_W$
& Stylus reference points used only for local HMD depth correction in $W$. \\

$\Gamma_S,\Gamma_Z,\Gamma_W$
& Specimen resection boundary and high-saliency depth-rim candidate sets in
frames $S$, $Z$, $W$. \\

\rowcolor{gray!7}
$\Tmat_{Z\leftarrow S}^{(0)}$
& Initial bead-based rigid transform from specimen frame $S$ to external
depth frame $Z$. \\

$\phi_Z(\cdot;\Theta),\ \Tmat_{\Delta}^{Z}$
& Contour-constrained deformation in $Z$ and the residual rigid correction
acting on deformed specimen points in $Z$. \\

\rowcolor{gray!7}
$\psi_{Z\leftarrow S},\ \Smat_{W\leftarrow Z}$
& Complete specimen-to-depth map and online marker-free depth-to-HMD fusion
transform. \\

\bottomrule
\end{tabularx}
\end{table*}

\subsection{Contour-constrained deformation and external depth bed registration}
\label{sec:deformation}

The resected specimen changes shape after excision because of tissue release, gravity, dehydration, and handling. We build on our prior Kelvinlet-based deformable registration framework for head and neck tumor resection, which modeled deformation with closed-form linear-elastic displacement bases and a uniform scale term for shrinkage \citep{Yang2025Deformable,pereira2026resection}. The present method adds an
explicit contour constraint so that the entire resection boundary, rather
than only four bead locations, contributes to the solution.

\subsubsection{Input surfaces and initialization}
\label{subsubsec:deform_inputs}

\begin{figure*}[t]
  \centering
  \includegraphics[width=\textwidth]{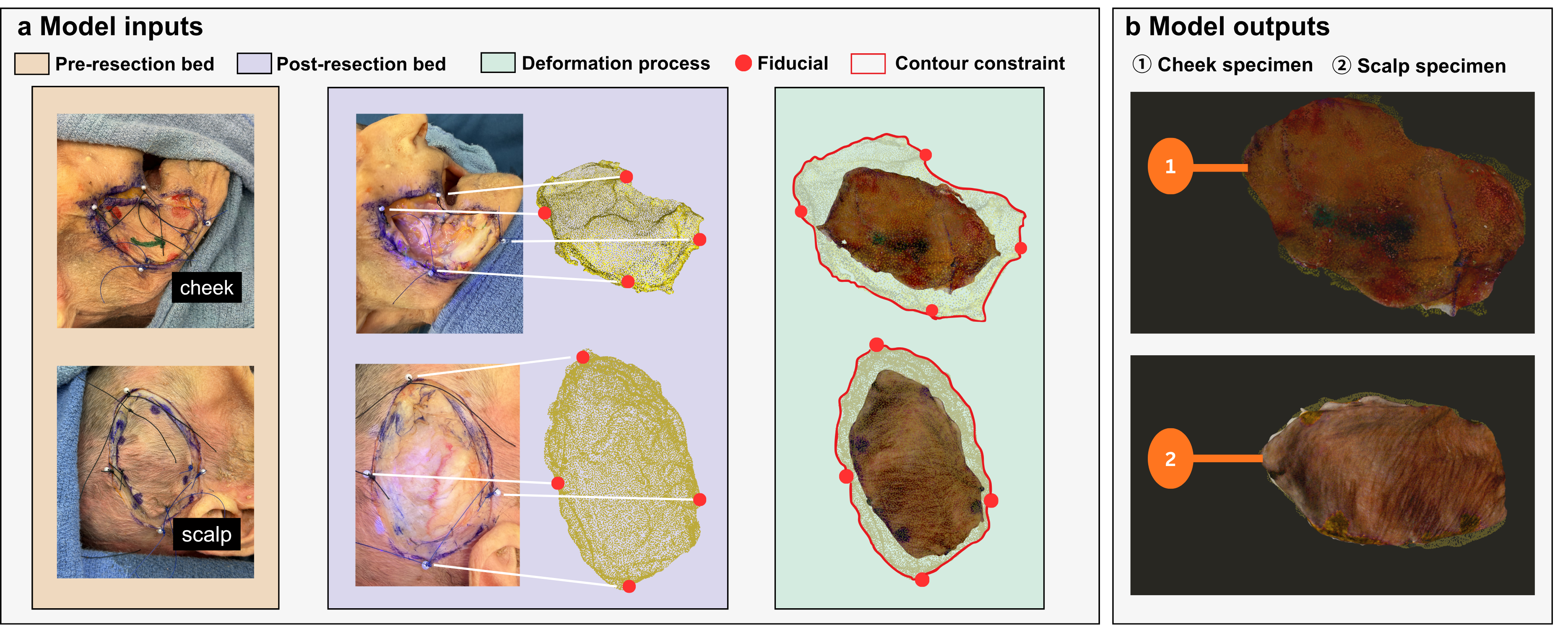}
  \caption{\textbf{(a)} Inputs for the contour-constrained deformation.
  \textbf{(b)} Target model outputs.}
  \label{fig:deform_register}
\end{figure*}

The post-resection bed geometry is acquired with the selected external depth
camera and processed to form the point cloud $\mathcal{B}_Z$. The bed is
segmented using Segment Anything Model 2 (SAM2) followed by manual
refinement. The patient-side resection boundary is identified from this
segmentation, and the boundary samples used by the deformation constraint and subsequent fusion are denoted by $\Gamma_Z$. The four patient-side bead
locations $\{\mathbf{a}_k^Z\}_{k=1}^{4}$ are also identified in frame $Z$.

The excised specimen is digitized with a structured-light scanner (EinScan
SP, Shining 3D, Hangzhou, China), producing the textured mesh
$\mathcal{M}_S$. The specimen resection boundary $\Gamma_S$, specimen-side
bead locations $\{\mathbf{a}_k^S\}_{k=1}^{4}$, and target labels are manually annotated in CloudCompare \citep{girardeau2016cloudcompare}. A tetrahedral volume is generated from $\mathcal{M}_S$, and 45 Kelvinlet control points are distributed through the volume.

The bead correspondences define the initial rigid transform
\begin{equation}
\Tmat_{Z\leftarrow S}^{(0)}=
\argmin_{\Tmat\in\SE(3)}
\sum_{k=1}^{4}
\left\|\mathbf{a}_k^Z-\Tmat\apply\mathbf{a}_k^S\right\|_2^2.
\label{eq:initial_rigid}
\end{equation}
This transform initializes the specimen in the external depth frame $Z$;
therefore, the common coordinate system for all subsequent deformation terms
is explicitly $Z$. The initialized specimen surface, $\mathcal{B}_Z$, the
bead correspondences, and the boundary sets $\Gamma_S$ and $\Gamma_Z$ are
then used for nonrigid estimation.

\subsubsection{Kelvinlet deformation with a contour constraint}
\label{subsubsec:kelvinlet_contour}

Nonrigid registration is performed using our previously published
contour-constrained Kelvinlet-based deformable registration framework
\citep{2606.19767}. In this approach, the deformation field
$\phi_Z(\cdot;\Theta)$ is modeled as a weighted combination of Kelvinlet
displacement modes generated from the selected control points while rigid,
isotropic-scale, and local deformation parameters are estimated jointly.

The framework incorporates three correspondence classes: bead landmarks,
surface points, and contour boundary points. The contour constraint uses the
full specimen resection boundary and corresponding cavity boundary, providing dense geometric information along the clinically relevant margin in addition to the sparse bead correspondences. Correspondence weights were selected based on the previous study, with bead and surface correspondences weighted at 1.0, contour correspondences weighted at 10.0, and a strain-energy regularization term weighted at $10^{-9}$.

The objective function is minimized using a Levenberg–Marquardt nonlinear least-squares optimizer, which estimates the contribution of each Kelvinlet control point to produce a smooth deformation field. The resulting deformation simultaneously aligns fiducial, surface, and contour features while minimizing strain energy. The optimized parameters $\Theta^{\star}$ are applied vertexwise to the initialized specimen mesh and to all specimen-side labels,
\begin{equation}
\mathcal{M}_Z^{\phi}=
\phi_Z\!\left(
\Tmat_{Z\leftarrow S}^{(0)}\apply\mathcal{M}_S;
\Theta^{\star}\right).
\label{eq:deformed_mesh}
\end{equation}
The resulting mesh $\mathcal{M}_Z^{\phi}$, deformed bead locations, and deformed targets are all expressed in frame $Z$ and are passed to the residual registration stage; they are not mapped to $W$ until the online AR fusion is applied. \cref{fig:deform_register} summarizes the model inputs and outputs.

\subsubsection{Residual registration to the resection bed}
\label{subsubsec:residual_registration}

After contour-constrained deformation, the specimen is already expressed in
the external depth frame, but a coherent pose bias can remain because of boundary extraction noise and scanner-to-external-depth bias. Such a bias can shift the transferred deep-margin labels even when the specimen boundary appears well aligned. Because the deep specimen surface and resection bed may overlap only partially and may provide weak local geometry, a direct local ICP update can be sensitive to initialization. We therefore estimate a bounded residual rigid correction before robust point-to-plane refinement.

Let
$\mathcal{X}_{\phi}^Z=\{\xvec_i^{\phi,Z}\}_{i=1}^{N_{\phi}}$
be samples from the deep surface of $\mathcal{M}_Z^{\phi}$. A
branch-and-bound estimate
$\Tmat_{\Delta,\mathrm{BnB}}^{Z}\in\SE(3)$ is obtained within the bounded
residual domain $\mathcal{D}_{\Delta}$,
\begin{align}
\Tmat_{\Delta,\mathrm{BnB}}^{Z}
&=\argmin_{\Tmat\in\mathcal{D}_{\Delta}}F_{\Delta}(\Tmat), \notag\\
F_{\Delta}(\Tmat)
&=\frac{1}{N_{\phi}}\sum_{i=1}^{N_{\phi}}
\min\!\left(d_i(\Tmat)^2,\tau_D^2\right), \notag\\
d_i(\Tmat)
&=\dist\!\left(
\Tmat\apply\xvec_i^{\phi,Z},\mathcal{B}_Z
\right).
\label{eq:residual_global_objective}
\end{align}
The truncation distance $\tau_D$ limits the influence of outliers and
non-overlap. The objective is a truncated ICP distance
\citep{Besl1992ICP} and is solved during the frozen section analysis waiting
interval using branch-and-bound search over $\mathcal{D}_{\Delta}$,
following the globally optimal ICP principle \citep{Yang2016GoICP}. For a
search cell centered at $\Tmat_c=(\Rmat_c,\tvec_c)$ with angular radius
$\theta_c$ and translation half-width $r_t$, the maximum displacement of
source point $i$ is bounded by
\begin{equation}
\epsilon_i=
2\sin\!\left(\frac{\theta_c}{2}\right)
\norm{\xvec_i^{\phi,Z}}+\sqrt{3}\,r_t.
\label{eq:bnb_radius}
\end{equation}
A valid lower bound for the cell is
\begin{align}
\underline{F}
&=\frac{1}{N_{\phi}}\sum_{i=1}^{N_{\phi}}
\min\!\left(
\max\{0,d_i^c-\epsilon_i\}^2,\tau_D^2
\right), \notag\\
d_i^c
&=\dist\!\left(
\Tmat_c\apply\xvec_i^{\phi,Z},\mathcal{B}_Z
\right).
\label{eq:bnb_lower}
\end{align}
The upper bound is obtained by evaluating
\cref{eq:residual_global_objective} at the cell center and by running short
local refinements from promising cells. The search terminates when the active
lower bound and the best upper bound differ by less than
$\varepsilon_{\mathrm{bnb}}$. Thus, the branch-and-bound estimate is globally
optimal up to $\varepsilon_{\mathrm{bnb}}$ within the specified residual pose
box.

The certified pose initializes a robust point-to-plane ICP refinement
\citep{chen1992object}. At iteration $r$,
\begin{equation}
\zvec_i^r=\Tmat^r\apply\xvec_i^{\phi,Z},
\qquad
\Tmat^0=\Tmat_{\Delta,\mathrm{BnB}}^{Z}.
\end{equation}
The fixed correspondence and target normal are
$\pvec_i^r=\Pi_{\mathcal{B}_Z}(\zvec_i^r)$ and
$\nvec_i^r=\nvec_{\pvec_i^r}$. Valid correspondences are collected in
$\mathcal{I}_r$ after distance and normal-compatibility filtering. The update
solves
\begin{align}
\Tmat^{r+1}
&=\argmin_{\Tmat\in\SE(3)}J_Z^r(\Tmat), \notag\\
J_Z^r(\Tmat)
&=\sum_{i\in\mathcal{I}_r}\eta_i^r\,
\rhoHuber\!\left(
(\nvec_i^r)^\top
[\Tmat\apply\xvec_i^{\phi,Z}-\pvec_i^r]
\right).
\label{eq:local_icp_zed}
\end{align}
Here, $\rhoHuber(\cdot)$ is the Huber robust loss with threshold $\delta$.
The nonnegative confidence weight $\eta_i^r$ increases with local source
curvature and decreases with target depth variance, emphasizing the
resection rim and high-confidence bed points. Let $r_\star$ denote the final
refinement iteration and define
$\Tmat_{\Delta}^{Z,\star}=\Tmat^{r_\star}$. The complete
specimen-to-external-depth map is then
\begin{equation}
\psi_{Z\leftarrow S}(\xvec^S)=
\Tmat_{\Delta}^{Z,\star}\apply
\phi_Z\!\left(
\Tmat_{Z\leftarrow S}^{(0)}\apply\xvec^S;
\Theta^\star
\right).
\label{eq:complete_specimen_depth_map}
\end{equation}

\subsection{Marker-free AR fusion}
\label{subsec:arfusion}

The AR fusion stage estimates $\Smat_{W\leftarrow Z}$, which maps the external depth-camera resection-bed point cloud and the deformed specimen model into the HMD world frame. This stage runs online while the surgeon looks at the bed. It therefore does not repeat the offline branch-and-bound residual registration from \cref{subsubsec:residual_registration}. Instead, it uses a design bounded by the surgical target size, precomputed external depth geometry, and a small number of local refinement iterations. The method is tailored to the HMD depth sensor and to draped surgical scenes in which flat drapes can dominate the point cloud.

\subsubsection{AHAT calibration and depth unprojection}
\label{subsubsec:hmd_calibration}

The HoloLens 2 HMD provides depth through the AHAT time-of-flight (ToF)
sensor. We use Research Mode to access AHAT depth and reflectivity frames
\citep{MicrosoftResearchMode2020}. The AHAT intrinsic parameters are
calibrated with reflectivity images of a planar target. Let
$\mathcal{C}=\{\mathbf{X}_k^R\}_{k=1}^{N_c}$ be target control points in the
calibration-target frame $R$, and let $\mathbf{u}_{jk}$ be their detected
pixel positions in frame $j$. The camera matrix $\mathbf{K}_H$, distortion
coefficients $\mathbf{d}_H$, and target-to-AHAT poses
$\Tmat_{H\leftarrow R}^{(j)}$ are estimated by
\begin{equation}
\argmin_{\mathbf{K}_H,\mathbf{d}_H,
\{\Tmat_{H\leftarrow R}^{(j)}\}}
\sum_{j,k}
\norm{
\mathbf{u}_{jk}-
\pi\!\left(
\mathbf{K}_H,\mathbf{d}_H,
\Tmat_{H\leftarrow R}^{(j)}\apply\mathbf{X}_k^R
\right)
}^2.
\label{eq:ahat_calibration}
\end{equation}
Here, $\pi(\cdot)$ is the distorted projection model. This follows standard
camera calibration \citep{Zhang2000Calibration} but uses AHAT reflectivity
rather than the front red-green-blue camera.

Each valid depth pixel $\mathbf{u}=[u,v,1]^\top$ with depth $z$ is
unprojected to AHAT coordinates,
\begin{equation}
\pvec^H=
z\,\pi^{-1}(\mathbf{u};\mathbf{K}_H,\mathbf{d}_H).
\label{eq:unprojection}
\end{equation}

Let $\mathcal{P}_{H,f}$ denote the set of valid AHAT points unprojected from
frame $f$. The HMD supplies the AHAT-to-world pose
$\Tmat_{W\leftarrow H}(t_f)$. Each frame is transformed to the HMD world
frame to form an uncorrected integrated HMD cloud,
\begin{equation}
\mathcal{P}_W^{0}=
\bigcup_{f=1}^{N_f}
\Tmat_{W\leftarrow H}(t_f)\apply\mathcal{P}_{H,f}.
\label{eq:hmd_integration}
\end{equation}
After the local depth correction described below, the corrected cloud is
cropped to the surgical working volume and denoted by $\mathcal{B}_W$.

\subsubsection{Region-specific HMD depth correction}
\label{subsubsec:depth_correction}

HoloLens 2 depth can support markerless surgical registration, but systematic depth errors can degrade patient-to-model alignment
\citep{kerkhof2025depth}. We correct the AHAT point cloud within the surgical target region rather than applying a global device correction. The local HMD depth-correction set
$\mathcal{L}_W=\{\boldsymbol{\ell}_i^W\}_{i=1}^{N_l}$ contains stylus
reference points in the predefined surgical working volume. It is used only
to estimate the HMD depth correction and does not contribute patient or
specimen correspondences to online fusion. Let $\pvec_i^{W,0}$ be the nearest uncorrected HMD depth point to $\boldsymbol{\ell}_i^W$. The correction transform is
\begin{equation}
\Tmat_c^W=
\argmin_{\Tmat\in\SE(3)}
\sum_{i=1}^{N_l}\beta_i
\left\|
\boldsymbol{\ell}_i^W-\Tmat\apply\pvec_i^{W,0}
\right\|^2.
\label{eq:depth_correction}
\end{equation}
The nonnegative weights $\beta_i$ down-weight noisy point pairs. The weighted Kabsch estimator solves
\cref{eq:depth_correction} by subtracting weighted centroids, forming a
covariance matrix, and applying singular value decomposition
\citep{Umeyama1991LeastSquares}. If the resulting rotation has negative
determinant, the final singular vector is flipped before the rotation is
recomputed. The corrected world point and corrected integrated cloud are
\begin{equation}
\widehat{\pvec}^{W}
=\Tmat_c^W\apply\pvec^{W,0},
\qquad
\widehat{\mathcal{P}}_W
=\Tmat_c^W\apply\mathcal{P}_W^0.
\label{eq:corrected_hmd_cloud}
\end{equation}

\subsubsection{Drape-aware saliency for surgical bed points}
\label{subsubsec:drape_saliency}

As shown in \cref{fig:ar_fusion_projection}a, the HMD depth cloud often
contains more drape area than resection area. Standard point-cloud
registration can therefore align the flat drape well and still miss the
smaller resection bed. We reduce this bias by estimating a drape likelihood
and a surgical saliency weight for every point. The following quantities are
computed independently for $\mathcal{B}_Z$ and $\mathcal{B}_W$; frame
superscripts are omitted for readability.

For point $\pvec_i$, let $\mathcal{N}_i$ be its spatial neighborhood and
\[
\bar{\pvec}_i=
|\mathcal{N}_i|^{-1}
\sum_{\pvec_j\in\mathcal{N}_i}\pvec_j
\]
its neighborhood centroid. The local covariance is
\begin{equation}
\mathbf{C}_i=
\frac{1}{|\mathcal{N}_i|}
\sum_{\pvec_j\in\mathcal{N}_i}
(\pvec_j-\bar{\pvec}_i)
(\pvec_j-\bar{\pvec}_i)^\top.
\label{eq:covariance}
\end{equation}

\par\medskip
\noindent
\begin{minipage}{\linewidth}
    \centering
    \includegraphics[width=\linewidth]
    {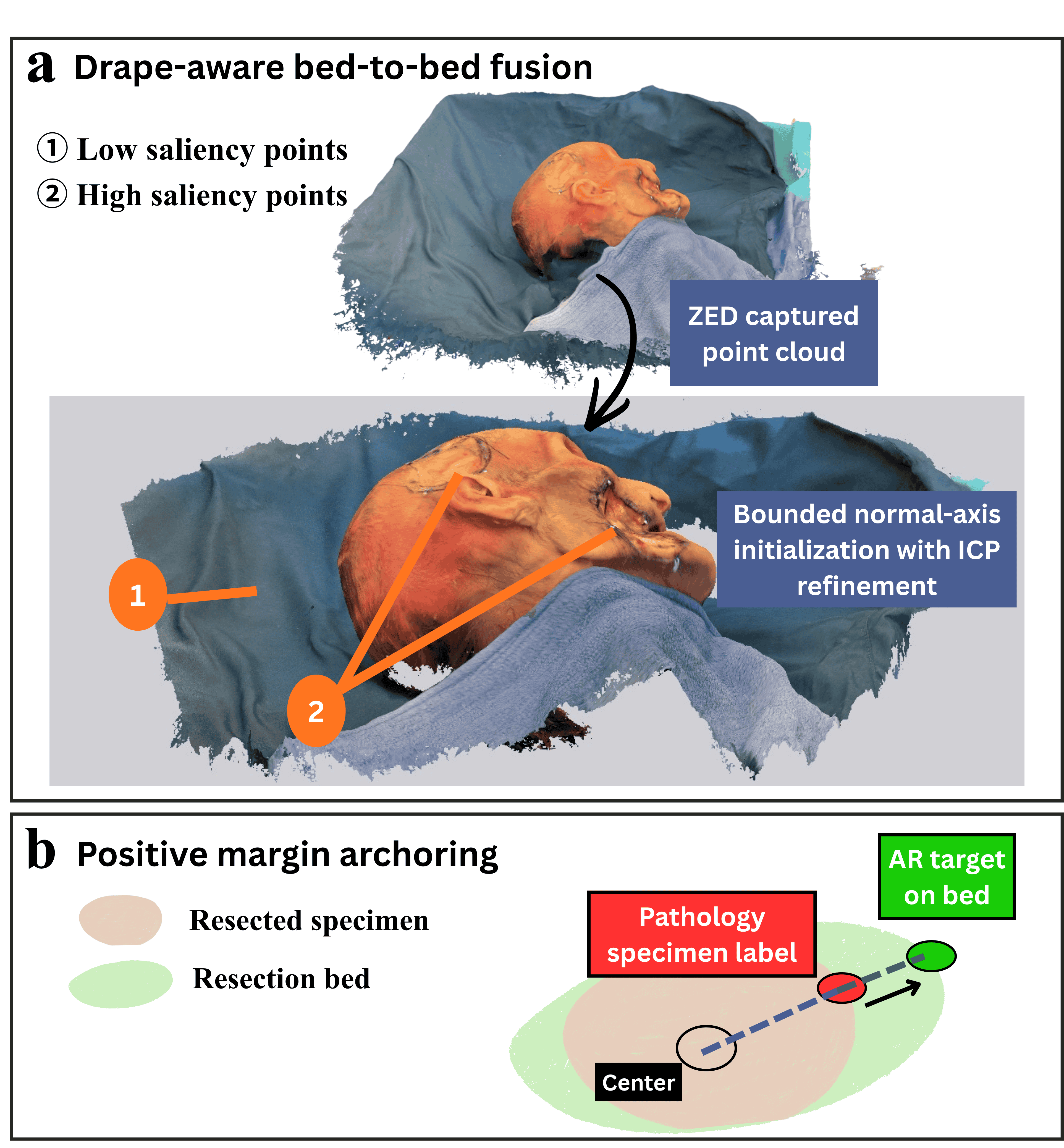}
    \captionof{figure}{Drape-aware AR fusion and margin anchoring.
    \textbf{(a)} Saliency weighting suppresses planar drape regions and
    emphasizes the resection rim and depth discontinuities.
    \textbf{(b)} After external depth-to-HMD fusion, each pathology label is
    projected so that the green target sphere lies on the reconstructed
    resection bed.}
    \label{fig:ar_fusion_projection}
\end{minipage}
\par\medskip

Let
$0\leq\lambda_{i1}\leq\lambda_{i2}\leq\lambda_{i3}$
be the eigenvalues of $\mathbf{C}_i$. We compute curvature and normal
variation as
\begin{equation}
\kappa_i=
\frac{\lambda_{i1}}
{\lambda_{i1}+\lambda_{i2}+\lambda_{i3}+\epsilon_\lambda},
\qquad
\nu_i=
1-
\norm{
\frac{1}{|\mathcal{N}_i|}
\sum_{\pvec_j\in\mathcal{N}_i}\nvec_j
}.
\label{eq:curvature_variation}
\end{equation}

For each cloud, a dominant drape plane
$\Pi=(\nvec_{\Pi},h_{\Pi})$ is estimated from low-curvature points using
random sample consensus (RANSAC) \citep{Fischler1981RANSAC}. Here,
$\nvec_{\Pi}$ is the unit plane normal and $h_{\Pi}$ is the scalar offset, so points on the plane satisfy
$\nvec_{\Pi}^{\top}\pvec+h_{\Pi}=0$. The signed plane distance is
$d_{\Pi}(\pvec)=\nvec_{\Pi}^{\top}\pvec+h_{\Pi}$. The drape likelihood is
\begin{equation}
\ell_i=
\exp\!\left[
-\frac{d_{\Pi}(\pvec_i)^2}{2\sigma_{\Pi}^2}
\right]
\exp\!\left[
-\frac{\kappa_i^2}{2\sigma_{\kappa}^2}
\right].
\label{eq:drape_likelihood}
\end{equation}
The surgical saliency weight is
\begin{align}
r_i
&=\eta\widetilde{\kappa}_i
 +(1-\eta)\widetilde{\nu}_i, \notag\\
d_i^{\perp}
&=\min\!\left(
1,\frac{|d_{\Pi}(\pvec_i)|}{\tau_{\perp}}
\right), \notag\\
w_i
&=w_{\min}+(1-w_{\min})
\left[
(1-\ell_i)r_i+\omega_{\perp}d_i^{\perp}
\right].
\label{eq:saliency_weight}
\end{align}
Here, $\widetilde{\kappa}_i$ and $\widetilde{\nu}_i$ are robustly normalized
to $[0,1]$; $\eta\in[0,1]$ balances normalized curvature and normal
variation; $\tau_{\perp}>0$ is the plane distance at which the off-plane
contribution saturates; $\omega_{\perp}\geq0$ weights that contribution; and
$w_{\min}>0$ prevents zero-weight regions. The scale parameters
$\sigma_{\Pi}$ and $\sigma_{\kappa}$ control the two factors in the drape
likelihood. The plane-distance term preserves depressed resection points that may be locally smooth.

\begin{algorithm*}[t]
\footnotesize
\caption{Runtime-bounded drape-aware AR fusion}
\label{alg:arfusion}
\begin{algorithmic}[1]
\Require Preprocessed external depth bed $\mathcal{B}_Z$, external depth rim
$\Gamma_Z$, HMD AHAT depth frames $\{D_f,t_f\}$, AHAT calibration
$(\mathbf{K}_H,\mathbf{d}_H)$, and local depth correction $\Tmat_c^W$
\Ensure Fusion transform $\Smat_{W\leftarrow Z}^{\star}$ and fusion quality
flag $q_F$

\State Unproject the AHAT frames using \cref{eq:unprojection}, transform them to $W$, and apply the correction in \cref{eq:corrected_hmd_cloud}

\State Crop to the surgical working volume and voxel-downsample to at most
$N_{\max}$ points

\If{the cropped HMD cloud has fewer than $N_{\min}$ points}
  \State Set $q_F\leftarrow$ recapture required and return without rendering
  guidance
\EndIf

\State Estimate normals, local covariance, curvature, and normal variation
for $\mathcal{B}_W$

\If{RANSAC finds a dominant drape plane with sufficient inlier support}
  \State Compute drape likelihoods using \cref{eq:drape_likelihood}
\Else
  \State Set $\ell_i\leftarrow0$ and rely on curvature and normal-variation
  weights
\EndIf

\State Compute saliency weights for $\mathcal{B}_W$ using
\cref{eq:saliency_weight}

\State Extract HMD rim candidates $\Gamma_W$ from high-saliency depth
discontinuities

\If{$\Gamma_W$ is incomplete}
  \State Set $\omega_{\Gamma}$ to its predefined reduced value and rely more
  strongly on the surface terms
\EndIf

\State Generate four normal-and-axis initialization candidates using
\cref{eq:tangent_bases,eq:candidate_rotations}

\For{each candidate $m$}
  \State Compute the weighted symmetric surface score and inlier ratio
  \If{the inlier ratio is below $r_{\min}$}
    \State Reject candidate $m$
  \EndIf
\EndFor

\If{all candidates are rejected}
  \State Set $q_F\leftarrow$ recapture required and return without rendering
  guidance
\Else
  \State Initialize \cref{eq:online_refine} with the best accepted candidate
\EndIf

\For{iteration $r=1,\ldots,R_F$}
  \State Update nearest-neighbor correspondences using distance, normal, and
  saliency gates
  \State Solve the bounded similarity point-to-plane update in
  \cref{eq:online_refine}
  \If{the change in transform is below $\varepsilon_F$}
    \State \textbf{break}
  \EndIf
\EndFor

\If{the final symmetric surface error and inlier ratio satisfy the acceptance
thresholds}
  \State Set $q_F\leftarrow$ accepted and return
  $\Smat_{W\leftarrow Z}^{\star}$
\Else
  \State Set $q_F\leftarrow$ low confidence and display a warning with the
  overlay
\EndIf
\end{algorithmic}
\end{algorithm*}

\subsubsection{Runtime-bounded external depth-to-HMD bed fusion}
\label{subsubsec:fast_fusion}

The external depth geometry is preprocessed during the pathology waiting
interval. We voxel-downsample $\mathcal{B}_Z$, estimate normals and saliency
weights, and retain $\Gamma_Z$ as the high-saliency depth-discontinuity points adjacent to the segmented resection boundary. A k-d tree over
$\mathcal{B}_Z$ supports HMD-to-external nearest-neighbor queries used in the symmetric initialization score. For each online capture, a second k-d tree is constructed over the downsampled $\mathcal{B}_W$ to evaluate the
external-to-HMD projections used during local refinement. The HMD cloud is
limited to $N_{\max}=8000$ high-saliency points so that these cross-cloud
queries and the Gauss--Newton updates remain bounded.

A fast initialization is generated from the surgical-bed geometry. Let
$\boldsymbol{\mu}_{\Gamma_Z}$ and $\boldsymbol{\mu}_{\Gamma_W}$ be the
3D saliency-weighted centroids of the two rim sets, and let
$\mathbf{C}_{\Gamma_Z}$ and $\mathbf{C}_{\Gamma_W}$ be their saliency-weighted covariance matrices. Let $\nvec_Z$ and $\nvec_W$ be the robust bed or drape-plane normals, oriented toward the camera. Principal rim axes are computed in the tangent planes orthogonal to these normals, giving the orthonormal bases
\begin{equation}
\mathbf{E}_Z=
[\mathbf{e}_{Z1},\mathbf{e}_{Z2},\nvec_Z],
\qquad
\mathbf{E}_W=
[\mathbf{e}_{W1},\mathbf{e}_{W2},\nvec_W].
\label{eq:tangent_bases}
\end{equation}
Because an ellipse-like rim can have sign ambiguity, we form four candidate
rotations
\begin{equation}
\Rmat_m=
\mathbf{E}_W\mathbf{Q}_m\mathbf{E}_Z^\top,
\qquad
\det(\Rmat_m)=1,
\label{eq:candidate_rotations}
\end{equation}
where $\mathbf{Q}_m$ contains the allowable sign flips in the tangent plane.
The bounded residual scale and corresponding translation are
\begin{align}
\alpha_m
&=\clip\!\left(
\sqrt{
\frac{\tr(\mathbf{C}_{\Gamma_W})}
{\tr(\mathbf{C}_{\Gamma_Z})}
},
1-\epsilon_{\alpha},
1+\epsilon_{\alpha}
\right), \notag\\
\tvec_m
&=\boldsymbol{\mu}_{\Gamma_W}
-\alpha_m\Rmat_m\boldsymbol{\mu}_{\Gamma_Z}.
\label{eq:candidate_scale_translation}
\end{align}
Here, $\clip(x,l,u)$ clamps $x$ to $[l,u]$. The scale interval is narrow and
compensates only residual depth bias after \cref{eq:depth_correction}; it does not change the physical interpretation of the specimen.

Each candidate $\Smat_m=(\alpha_m,\Rmat_m,\tvec_m)$ is scored with a saliency-weighted symmetric surface distance. The best accepted candidate initializes the local refinement. For $\Smat=(\alpha,\Rmat,\tvec)$,
\begin{align}
\Smat_{W\leftarrow Z}^{\star}
&=\argmin_{\substack{
\Smat=(\alpha,\Rmat,\tvec)\in\Sim(3)\\
\alpha\in[1-\epsilon_{\alpha},1+\epsilon_{\alpha}]
}}
J_F(\Smat), \notag\\
J_F(\Smat)
&=\sum_{i\in\mathcal{I}}
w_i^Z w_{\pi_i}^W
\rhoHuber\!\left(
(\nvec_{\pi_i}^W)^\top
\mathbf{e}_i(\Smat)
\right) \notag\\
&\quad+
\omega_{\Gamma}
\sum_{\mathbf{g}\in\Gamma_Z}
\rhoHuber\!\left(
\dist(
\Smat\apply\mathbf{g},
\Gamma_W
)
\right)
+\omega_{\alpha}(\alpha-1)^2.
\label{eq:online_refine}
\end{align}
where
\begin{equation}
\qvec_{\pi_i}^W=
\Pi_{\mathcal{B}_W}
(\Smat\apply\pvec_i^Z),
\qquad
\mathbf{e}_i(\Smat)=
\Smat\apply\pvec_i^Z-\qvec_{\pi_i}^W.
\label{eq:online_correspondence}
\end{equation}
The nonnegative scalars $\omega_{\Gamma}$ and $\omega_{\alpha}$ are objective weights. The rim weight $\omega_{\Gamma}$ takes a predefined nominal or reduced value according to rim completeness. Correspondences are updated with capped iterations using distance, normal, and saliency gates. The final specimen-to-world map is
\begin{equation}
\psi_{W\leftarrow S}(\xvec^S)=
\Smat_{W\leftarrow Z}^{\star}\apply
\psi_{Z\leftarrow S}(\xvec^S).
\label{eq:specimen_to_world}
\end{equation}

The online computational cost is dominated by the k-d tree nearest-neighbor
queries and a small Gauss--Newton system. With precomputed external-depth
saliency, a voxel-downsampled HMD cloud, $N_{\max}$ bounded to 8000 points,
four initialization candidates, and fewer than 15 refinement iterations, no
dense feature matching is performed online.

\subsection{Textured rendering and positive margin anchoring}
\label{subsec:rendering}

After AR fusion, the deformed specimen mesh and positive-margin targets are
represented in the HMD world frame. The rendering module displays the
deformed specimen as a semi-transparent textured surface and the positive
margins as green spheres (see \cref{fig:workflow}a and
\cref{fig:eval_fusion}b). In this implementation, each target margin is
represented by a point rather than an area or interval. The transparency is
set to 50\% in the Unity Engine Standard Shader. This choice allows the
specimen shape to act as a visual registration check while preserving the
surgeon's view of the physical resection bed through the optical see-through
visor. The deformation updates mesh vertex positions but preserves mesh
connectivity and the original texture coordinates. The renderer therefore
applies $\psi_{W\leftarrow S}$ to the vertices of $\mathcal{M}_S$ and samples the original texture atlas at the unchanged coordinates; no separate
texture-registration step is required.

\begin{algorithm*}[t]
\footnotesize
\caption{Positive-margin anchoring on the resection bed}
\label{alg:margin_anchor}
\begin{algorithmic}[1]
\Require Pathology label $\mvec_j^S$, map $\psi_{Z\leftarrow S}$, external
depth bed $\mathcal{B}_Z$, optional bed mesh $\mathcal{S}_Z$, and fusion
transform $\Smat_{W\leftarrow Z}^{\star}$
\Ensure Rendered AR target center $\widetilde{\mvec}_j^W$

\State Transform the label to external depth coordinates using
\cref{eq:margin_zed}

\State Estimate the robust contour center $\cvec_{\Gamma}^{Z}$ using
\cref{eq:geomedian}

\State Compute the tangent-plane radial direction $\dvec_j^Z$ using
\cref{eq:ray_direction}

\If{a triangulated bed surface $\mathcal{S}_Z$ is available}
  \State Compute all valid positive ray--triangle intersections
  $\mathcal{R}_j$ using \cref{eq:valid_ray_intersections}
\Else
  \State Set $\mathcal{R}_j\leftarrow\emptyset$
\EndIf

\If{$\mathcal{R}_j\neq\emptyset$}
  \State Select the smallest positive ray parameter using
  \cref{eq:ray_intersection}
\Else
  \State Query bed points in a cylinder around the ray using
  \cref{eq:cyl_query}
  \If{$\mathcal{Q}_j\neq\emptyset$}
    \State Compute a ray-weighted bed point using
    \cref{eq:ray_weighted_point}
  \Else
    \State Use the closest point in $\mathcal{B}_Z$ to $\mvec_j^Z$ and flag
    the target as fallback-anchored
  \EndIf
\EndIf

\State Transform the anchored depth target to HMD world coordinates using
\cref{eq:margin_world}

\State \Return $\widetilde{\mvec}_j^W$
\end{algorithmic}
\end{algorithm*}

\subsubsection{Radial bed anchoring of positive-margin labels}
\label{subsubsec:margin_projection}

A point-valued pathology label is denoted by the bold lower-case vector
$\mvec_j^S$, whereas $\mathcal{M}_S$ denotes the complete specimen mesh.
Likewise, $\cvec_{\Gamma}^{Z}$ denotes the bold lower-case contour-center
vector. The pathologist labels a positive margin on the scanned specimen mesh as $\mvec_j^S$. As shown in \cref{fig:ar_fusion_projection}b, uncertainty or residual registration error can place $\psi_{W\leftarrow S}(\mvec_j^S)$ on a specimen surface that is slightly larger, smaller, or offset from the reconstructed patient bed. We therefore anchor every displayed target to $\mathcal{B}_Z$ before mapping it into the HMD world frame.

The label after deformation and residual depth registration is
\begin{equation}
\mvec_j^Z=
\psi_{Z\leftarrow S}(\mvec_j^S).
\label{eq:margin_zed}
\end{equation}
A robust contour center is estimated from the deformed, registered specimen
contour,
\begin{equation}
\cvec_{\Gamma}^{Z}=
\argmin_{\cvec\in\R^3}
\sum_{\mathbf{g}\in
\psi_{Z\leftarrow S}(\Gamma_S)}
\norm{\mathbf{g}-\cvec}.
\label{eq:geomedian}
\end{equation}
For round or elliptical cheek and scalp resections, shrinkage is expected to
occur predominantly toward the specimen center. We use this radial assumption to preserve angular direction while forcing the displayed target onto the depth-bed surface. Let $\nvec_B^Z$ be the robust bed normal near the resection, and let
$\mathbf{P}_B=\I-\nvec_B^Z\nvec_B^{Z\top}$ be the tangent-plane projector.
Define
\begin{equation}
\mathbf{r}_j^Z=
\mathbf{P}_B
(\mvec_j^Z-\cvec_{\Gamma}^{Z}).
\label{eq:radial_vector}
\end{equation}
If $\norm{\mathbf{r}_j^Z}<\epsilon_d$, the radial direction is treated as
degenerate and the method uses the closest-point fallback on
$\mathcal{B}_Z$. Otherwise,
\begin{equation}
\dvec_j^Z=
\frac{\mathbf{r}_j^Z}
{\norm{\mathbf{r}_j^Z}}.
\label{eq:ray_direction}
\end{equation}
This bed-plane projection makes the ray well defined for point clouds, prevents out-of-plane specimen error from determining the target direction,
and ensures that $\dvec_j^Z$ has unit length.

For a triangulated bed surface $\mathcal{S}_Z$, the algorithm tests the ray
against all triangles and retains the positive intersections whose locations
lie inside the segmented resection boundary,
\begin{equation}
\mathcal{R}_j=
\left\{
a>0\;\middle|\;
\cvec_{\Gamma}^{Z}+a\dvec_j^Z
\text{ intersects a triangle of }\mathcal{S}_Z
\right\}.
\label{eq:valid_ray_intersections}
\end{equation}
If $\mathcal{R}_j\neq\emptyset$, the nearest valid surface point is selected,
\begin{equation}
a_j=\min\mathcal{R}_j,
\qquad
\widetilde{\mvec}_j^Z=
\cvec_{\Gamma}^{Z}+a_j\dvec_j^Z.
\label{eq:ray_intersection}
\end{equation}
The result is therefore independent of triangle traversal order. If no valid
mesh intersection exists, the point-cloud procedure below is used.

For a point cloud, define
$\Pmat_j^{Z\perp}=\I-\dvec_j^Z\dvec_j^{Z\top}$. We query bed points in a
cylinder around the ray,
\begin{align}
\mathcal{Q}_j=
\{\pvec\in\mathcal{B}_Z\mid
&\norm{
\Pmat_j^{Z\perp}
(\pvec-\cvec_{\Gamma}^{Z})
}
\leq r_q, \notag\\
&\dvec_j^{Z\top}
(\pvec-\cvec_{\Gamma}^{Z})
>0
\}.
\label{eq:cyl_query}
\end{align}
The point-cloud target is a ray-weighted average,
\begin{align}
\widetilde{\mvec}_j^Z
&=
\frac{
\sum_{\pvec\in\mathcal{Q}_j}
\omega_{j\pvec}\pvec
}{
\sum_{\pvec\in\mathcal{Q}_j}
\omega_{j\pvec}
}, \notag\\
\omega_{j\pvec}
&=
\exp[
-r_{j\pvec}^2/(2\sigma_q^2)
], \notag\\
r_{j\pvec}
&=
\norm{
\Pmat_j^{Z\perp}
(\pvec-\cvec_{\Gamma}^{Z})
}.
\label{eq:ray_weighted_point}
\end{align}
If $\mathcal{Q}_j$ is empty, the method falls back to the closest point on
$\mathcal{B}_Z$ to $\mvec_j^Z$. The final AR sphere center is
\begin{equation}
\widetilde{\mvec}_j^W=
\Smat_{W\leftarrow Z}^{\star}\apply
\widetilde{\mvec}_j^Z.
\label{eq:margin_world}
\end{equation}
This anchoring step, described in \cref{alg:margin_anchor}, defines a
surgeon-facing target that is consistent with the pathologist's specimen
label, the radial shrinkage assumption, the deformed contour, and the
reconstructed depth resection bed.

\section{Formative Experiments}
We evaluated the proposed system in two stages. First, formative experiments were performed to configure the workflow, including depth-camera selection for resection-bed acquisition and validation of marker-free fusion relative to a marker-based reference. Second, evaluation experiments were performed, including a user study, to evaluate system efficacy. \Cref{fig:eval_fusion} summarizes the formative experiments and the user study protocol. In this section, we focus on the formative experiments.

\subsection{Depth-camera selection}
\label{subsec:depth_camera_selection}
We first compared two candidate patient-side depth cameras because the completeness of the resection-bed point cloud affects subsequent deformation, registration, and AR fusion: ZED (ZED 2i, Stereolabs Inc., USA) and Zivid (Zivid 2 M70, Zivid AS, Norway). The EinScan SP was not included because it served a different role in the workflow (close-range ex vivo specimen digitization) and was not a candidate for overhead patient-side resection-bed capture. The comparison was intended to select hardware for this workflow rather than to establish a general ranking of the devices.

Our camera comparison was performed on buccal targets, which provide a more challenging capture scenario than the cutaneous cheek and scalp beds because the oral cavity is more constrained and more susceptible to occlusion and reflective lighting. Before data collection, camera-specific exposure and depth-quality settings were adjusted under fixed lighting to maximize valid target coverage. The selected settings were then held constant across targets and distances. Let $c \in \{\mathrm{ZED}, \mathrm{Zivid}\}$ denote the camera, let $d$ denote the scan distance, and let $t$ denote the target. The CT-derived target surface is $\mathcal{R}_{d,t}$ and the segmented camera point cloud is $\mathcal{P}_{c,d,t}$. We measured target coverage as
\begin{equation}
C_{c,d,t}=100\frac{A\left(\{\mathbf{x}\in\mathcal{R}_{d,t}:\dist(\mathbf{x},\mathcal{P}_{c,d,t})<\tau_C\}\right)}{A(\mathcal{R}_{d,t})},
\end{equation}
where $A(\cdot)$ is surface area and $\tau_C$ is the capture tolerance. We also counted segmented target points,
\begin{equation}
n_{c,d,t}=|\mathcal{P}_{c,d,t}|,
\end{equation}
which reflects how much target geometry remains available for registration after cropping and segmentation.

\begin{figure*}[t]
\centering
\includegraphics[width=\textwidth]{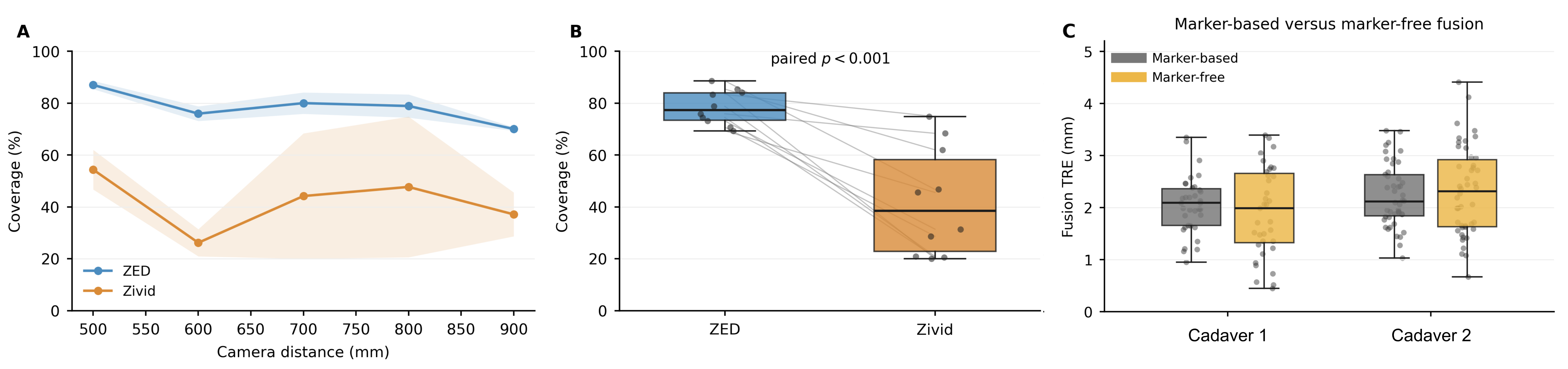}
\caption{Formative experiment results. \textbf{(a)} Target coverage from 500 to 900 mm; the shaded band denotes $\pm1$ standard deviation across repeated scans. \textbf{(b)} Paired coverage comparison between the two cameras. \textbf{(c)} Marker-based versus marker-free AR fusion error measured on two cadavers. }
\label{fig:formative}
\end{figure*}

As shown in \cref{fig:formative}a and b, the camera comparison showed a large coverage difference across the two buccal targets and five distances. Under the reported workflow settings, ZED achieved $78.4 \,{\pm}\, 6.7\%$ target coverage, whereas Zivid achieved $41.9 \,{\pm}\, 20.8\%$. The paired difference was 36.5 percentage points in favor of ZED ($p<0.001$). ZED also preserved more segmented target points on average ($3875 \,{\pm}\, 2355$ vs.\ $3143 \,{\pm}\, 2062$ points). Although Zivid occasionally captured one side well at longer distances, it lost larger regions under reflective surgical-light conditions. We therefore selected \ZED for all subsequent experiments. Throughout the remainder of this section, the external depth-camera frame corresponds to the \ZED frame.
\par\medskip
\noindent
\begin{minipage}{\linewidth}
\centering
\includegraphics[width=\linewidth]{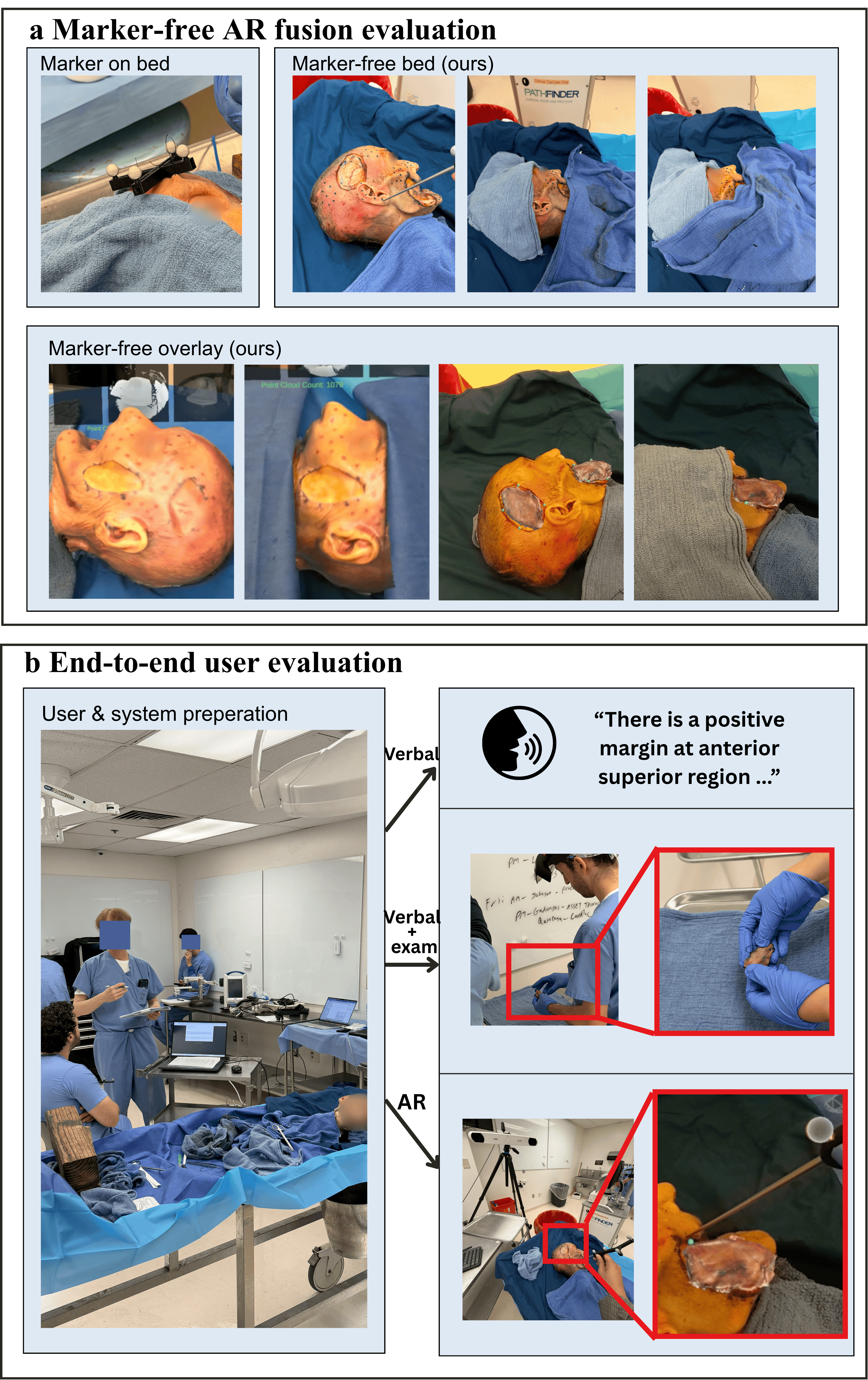}
\captionof{figure}{Evaluation protocols. \textbf{(a)} Formative evaluation of marker-free AR fusion against a marker-based reference and under different drape occlusion levels. \textbf{(b)} Cadaveric user study protocol for end-to-end positive-margin localization under verbal guidance, specimen examination, and AR guidance.}
\label{fig:eval_fusion}
\end{minipage}
\par\medskip

\subsection{Marker-free AR fusion compared with marker-based fusion}
Traditional optical see-through AR guidance often uses markers attached near the target anatomy (\cref{fig:eval_fusion}a). Our system instead estimates $\mathbf{S}_{W\leftarrow Z}$ from resection-bed geometry and HMD depth. Before the participant evaluation, we performed a formative cadaver study to compare this marker-free fusion with a marker-based reference workflow. A tracked stylus was used to point-match physical ink landmarks on the cadaver head. For each landmark $r$, the no-overlay measurement defined the physical reference $\mathbf{h}^{W}_{r}$, and the overlay-occluded measurement under fusion method $m$ defined $\widehat{\mathbf{h}}^{W}_{r,m}$. The AR fusion target registration error was
\begin{equation}
e^{F}_{r,m}=\left\|\widehat{\mathbf{h}}^{W}_{r,m}-\mathbf{h}^{W}_{r}\right\|_2.
\end{equation}
The first cadaver included $16$ nose, $11$ cheek, and $8$ jaw landmarks. The second cadaver, also used for the evaluation experiment, included $8$ nose, $8$ cheek, $8$ ear, $6$ jaw, and $16$ head landmarks. We compared marker-based and marker-free errors with paired $t$ tests because the same physical points were measured under both methods. Region-level differences were tested with Kruskal--Wallis tests. The stylus was tracked simultaneously by the HMD and the NDI Polaris, following prior work on tool tracking with off-the-shelf AR HMDs \citep{martingomez2024sttar}.

\cref{fig:formative}c shows the observed fusion errors. No significant paired difference was detected between marker-free and marker-based fusion on either cadaver. On the first cadaver, marker-based fusion measured $2.05 \pm 0.55\mm$ and marker-free fusion measured $1.92 \pm 0.85\mm$ ($p=0.359$). On the second cadaver, the corresponding values were $2.23 \pm 0.60\mm$ and $2.32 \pm 0.85\mm$ ($p=0.530$). The combined marker-free mean was $2.15 \pm 0.87\mm$, with a maximum of $4.41\mm$. Because the experiment was not designed as an equivalence or non-inferiority study, these findings indicate similar observed error magnitudes rather than proof that the two approaches are interchangeable. The marker-based workflow was more sensitive to incomplete capture of the retro-reflective spheres.

\section{Evaluation Experiments}
The insights and refinements detailed in the formative phase informed the finalized design of our system. To establish the efficacy of this final system, we conducted subsequent evaluation experiments, including a cadaveric user study, to evaluate component-level deformation, registration, and AR fusion accuracy, end-to-end margin localization accuracy, subjective user feedback, and latency on cheek and scalp resections. 

\subsection{Cadaveric user study design}
\input{tables/table_setup.tex}

As summarized in \cref{tab:setup}, we evaluated end-to-end positive-margin relocation on two cadaveric head and neck sites (\cref{fig:workflow}b): a cheek resection and a scalp resection. For each specimen, four peripheral targets and two deep targets were defined. Five participants (three male and two female) were recruited, with a mean age of $33\pm5$ years. Three participants were residents, one was a fellow, and one was an attending surgeon. Four participants were in otolaryngology--head and neck surgery, and one resident was in general surgery on a non-otolaryngology rotation. All participants reported some prior familiarity with VR and AR systems.

Before data collection, the experimenter provided a standardized training video featuring an experienced surgeon explaining the three guidance conditions and instructing each participant to indicate the perceived positive-margin location by placing the tracked stylus tip on
the resection bed. The same task description and condition-specific information were used for all participants based on our study protocol. As shown in \cref{fig:eval_fusion}b, each participant completed margin-relocation tasks under three guidance conditions. The verbal condition provided only a regional pathologic description of the positive margin. The examination condition allowed the participants to inspect the resected specimen and its orientation sutures, in addition to verbal instructions, before returning to the resection bed. These first two conditions represent current clinical practice, with the choice between them depending on surgeon preference and institutional resources. The AR condition displayed green target spheres through the HMD, and the peripheral-target tasks also showed the semi-transparent deformed specimen overlay.

To create blinded ground truth, the positive margins were marked with visible ink on the specimen for digital labeling, whereas the corresponding targets on the resection bed were marked with invisible ultraviolet (UV) ink. The visible specimen ink was removed after scanning, which prevented direct visual identification of the labeled location during the examination condition. Immediately after each localization task, the experimenter illuminated the bed with UV light and measured the hidden target location, as visualized in \cref{fig:uv}. Participants also completed NASA-Task Load Index (NASA-TLX) forms after localization tasks, System Usability Scale (SUS) forms after AR task blocks, and a questionnaire about the system after AR use (AR-specific). At the end of the study protocol, an experienced human factors specialist interviewed each participant to elicit their overall perspectives about the system and its performance.

\par\medskip
\noindent
\begin{minipage}{\linewidth}
\centering
\includegraphics[width=\linewidth]{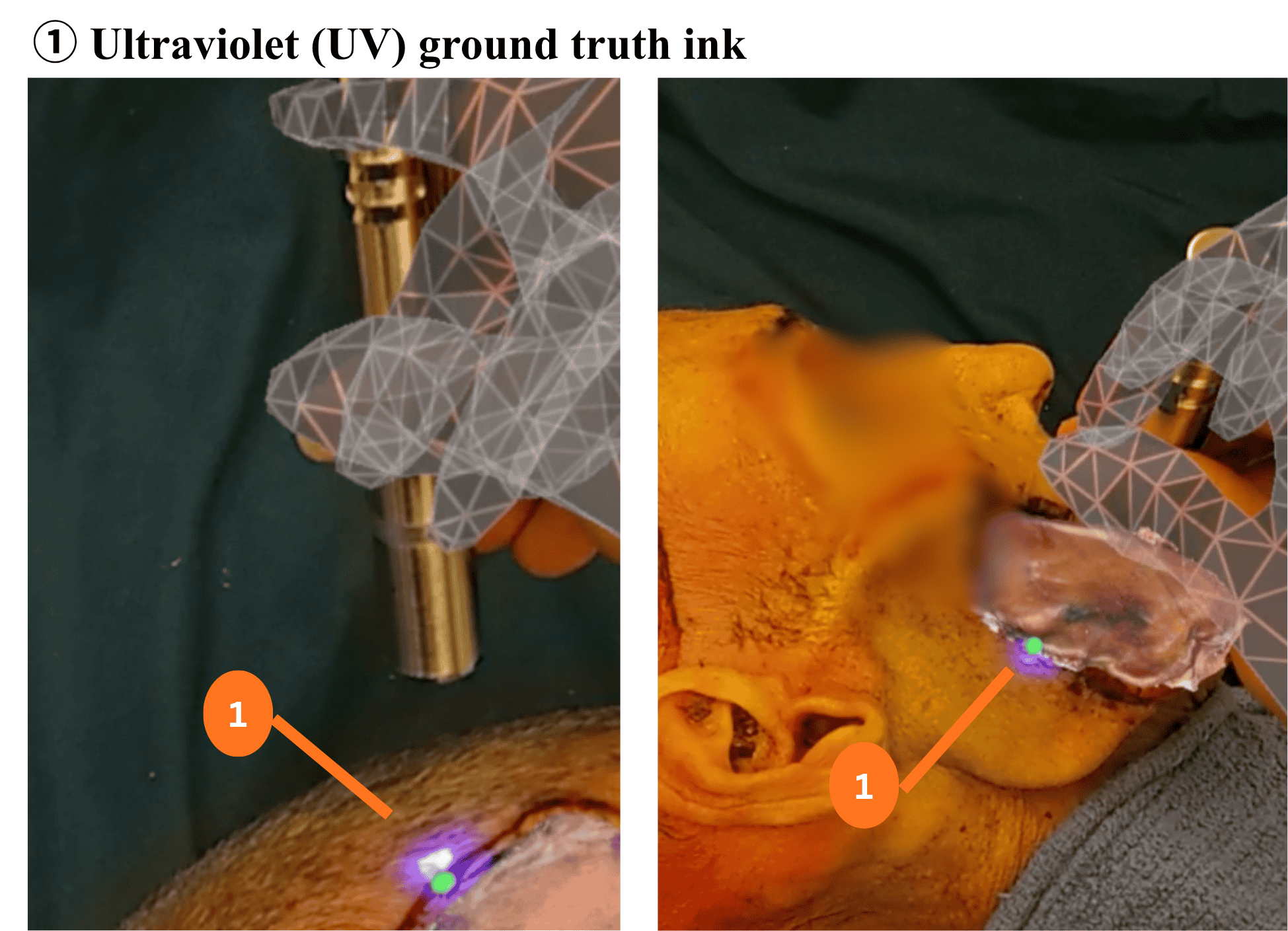}
\captionof{figure}{Visualization of the ground truth position with a UV light held by the experimenter. Margins are viewed through the AR HMD alongside our system's predicted position as green dot.}
\label{fig:uv}
\end{minipage}
\par\medskip

\subsection{Component-level accuracy on the user-study evaluation specimens}

\subsubsection{Deformation accuracy}
\label{subsec:deformation_metric}
Deformation accuracy was assessed using target registration error (TRE) computed from paired UV surface markers placed on both the resection bed and the specimen. Importantly, these UV markers were not used during registration and were distinct from the bead-suture correspondences used by the deformation algorithm described in \cref{sec:deformation}, which ensured an independent assessment of surface accuracy. TRE was calculated as the root-mean-square Euclidean distance between the corresponding target locations on the deformed specimen and the resection bed:
\begin{equation}
\label{TREequation}
\mathrm{TRE}=
\sqrt{\frac{1}{M}\sum_{i=1}^{M}\left\|p_i-q_i\right\|^2}
\end{equation}
where $M$ denotes the number of target points, $p_i$ denotes the location of the $i^{\mathrm{th}}$ target on the deformed specimen, and $q_i$ denotes the corresponding target location on the resection bed. The mean TRE across participants was $3.72\pm1.02\mm$ for the scalp specimen and $7.63\pm3.74\mm$ for the cheek specimen.

\begin{figure*}[t]
\centering
\includegraphics[width=\textwidth]{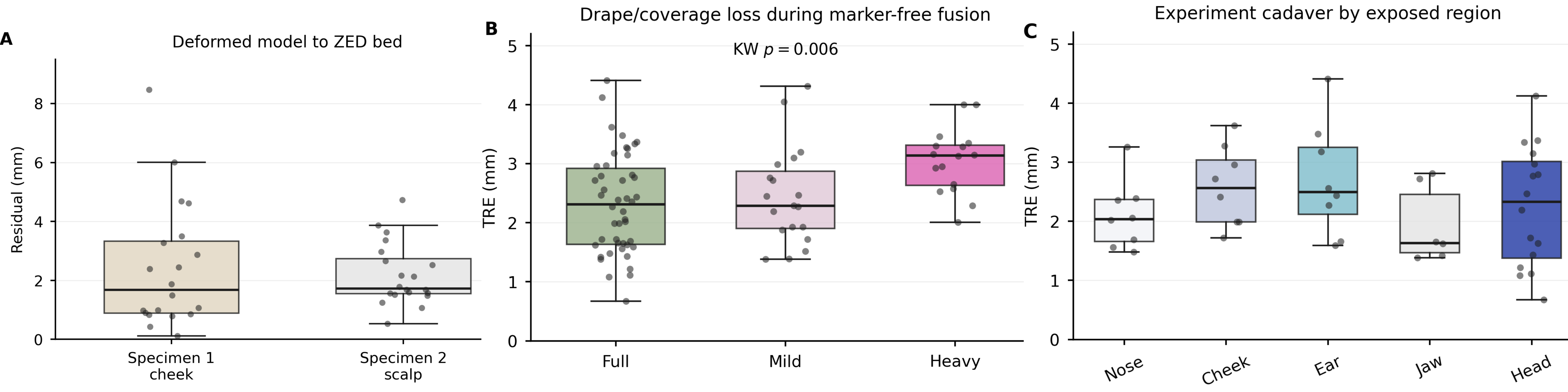}
\caption{Component-level accuracy. \textbf{(a)} Residual error after registering the deformed specimen to the \ZED resection bed. \textbf{(b)} Marker-free fusion error under full exposure, mild occlusion, and heavy occlusion. \textbf{(c)} Marker-free fusion error stratified by exposed facial anatomical region on the experiment cadaver.}
\label{fig:component_accuracy}
\end{figure*}

\subsubsection{Registration of the deformed specimen to the \ZED resection bed}

After contour-constrained deformation, the deformed model must align with the post-resection \ZED point cloud at the deep portion of the resection bed. This component isolates patient-side placement of the deformed specimen from HMD fusion and user interpretation. Let $\mathcal{Q}_{s}=\{\mathbf{q}^{S}_{s,j}\}_{j=1}^{N_s}$ be held-out deep-bed validation samples on specimen $s$ that were not used to fit the deformation. The sample after deformation and \ZED registration is $\psi_{Z\leftarrow S}(\mathbf{q}^{S}_{s,j})$. The bed registration residual was
\begin{equation}
e^{Z}_{s,j}=\dist\left(\psi_{Z\leftarrow S}(\mathbf{q}^{S}_{s,j}),\mathcal{B}^{Z}_{s}\right).
\label{eq:zed_bed_residual}
\end{equation}
The deformed specimen registered to the \ZED bed with millimeter-level residuals in both cases. The cheek specimen had a mean residual of $2.43\pm2.15\mm$, and the scalp specimen had a mean residual of $2.19\pm1.06\mm$. This measures the distance of held-out bed samples
to the observed \ZED bed surface after registration. The
approximately $3\mm$  residuals therefore indicate that the refinement placed the deformed specimen close to the patient-side bed geometry, even though local point-to-point correspondence errors from deformation could remain. The cheek--scalp comparison was not significant (Mann--Whitney, $p=0.499$). As shown in \cref{fig:component_accuracy}a, the cheek specimen had a wider spread and higher IQR than the scalp specimen.

\subsubsection{Marker-free fusion under different occlusion levels}

The comparison of AR fusion error in our formative study assumes that most of the head is visible. In the operating room, surgical drapes cover a large fraction of the head and leave only a limited surgical field exposed. To quantify this effect in our summative phase, we repeated the same stylus point-matching experiment under increasingly occlusive conditions. Let $o$ denote the occlusion level and let $\mathcal{R}_{o}$ be the set of exposed validation points. The occlusion-specific fusion error was
\begin{equation}
e^{O}_{r,o}=\left\|\widehat{\mathbf{h}}^{W}_{r,o}-\mathbf{h}^{W}_{r}\right\|_2,\qquad r\in\mathcal{R}_{o}.
\end{equation}

As shown in \cref{fig:component_accuracy}b, the error distributions in median [Q1, Q3] for full exposure, mild occlusion, and heavy occlusion were $2.31$ [1.63, 2.92], $2.29$ [1.90, 2.88], and $3.14$ [2.63, 3.31] respectively ($p=0.006$ with Kruskal--Wallis test), indicating that marker-free fusion accuracy was most affected under heavy occlusion. The right panels of \cref{fig:eval_fusion}a illustrate the extent of this heavy-occlusion condition. \cref{fig:component_accuracy}c further stratifies marker-free fusion errors by anatomical region under the full-exposure condition.

\subsection{End-to-end localization accuracy}

\begin{figure*}[t]
\centering
\includegraphics[width=0.98\textwidth]{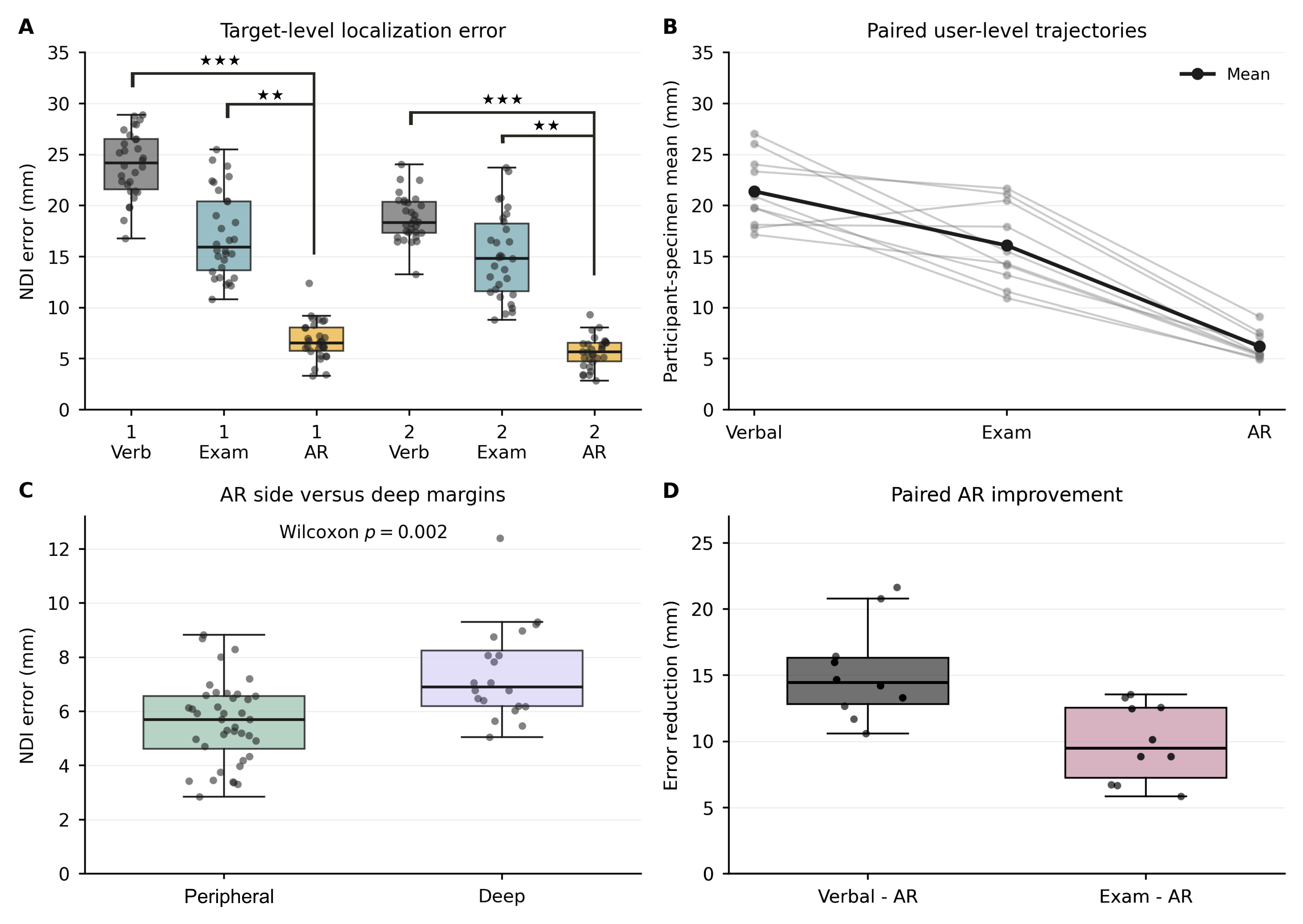}
\caption{End-to-end localization under NDI coordinates. Significance levels are directly labeled or indicated by asterisks: $^{*}p < 0.05$, $^{**}p < 0.01$, and $^{***}p < 0.001$. \textbf{(a)} Target-level errors for each specimen and condition. \textbf{(b)} Participant-specimen means. \textbf{(c)} Peripheral and deep AR targets shown separately under NDI stylus tracking. \textbf{(d)} Paired AR improvements.}
\label{fig:end_to_end_ndi}
\end{figure*}

The end-to-end task evaluated the clinically relevant question: whether a surgeon could localize the positive margin on the resection bed under each guidance condition. Let $i$ index participant, $s$ index specimen, $c$ index one of the three guidance conditions, and $j\in\{1,\ldots,6\}$ index target. The participant-selected stylus point is $\mathbf{r}^{k,W}_{i,s,c,j}$, and the UV ground-truth point is $\mathbf{g}^{k,W}_{s,j}$. The coordinate source $k$ is NDI Polaris or HoloLens inside-out tracking of the stylus. The end-to-end localization error was
\begin{equation}
e^{\mathrm{E2E},k}_{i,s,c,j}=\left\|\mathbf{r}^{k,W}_{i,s,c,j}-\mathbf{g}^{k,W}_{s,j}\right\|_2.
\end{equation}
For participant-level paired tests, we averaged the six targets for each participant, specimen, and condition,
\begin{equation}
\bar{e}^{\mathrm{E2E},k}_{i,s,c}=\frac{1}{6}\sum_{j=1}^{6} e^{\mathrm{E2E},k}_{i,s,c,j}.
\end{equation}
We used the Friedman test across the three guidance conditions, followed by Holm-corrected Wilcoxon signed-rank tests on participant-specimen means \citep{friedman1937rank,wilcoxon1945individual,holm1979simple}.

AR guidance reduced end-to-end localization error (\cref{fig:end_to_end_ndi}). Under NDI coordinates, relocation errors were $21.40\pm3.84\mm$ under verbal guidance across both specimens, $16.09\pm4.30\mm$ under specimen examination, and $6.19\pm1.79\mm$ under AR guidance. AR-guided localization resulted in a 70\% decrease in the error from the current standard of care. The cheek specimen had a slightly higher AR median and a wider AR IQR than the scalp specimen ($6.56$ [5.76, 8.05] mm vs.\ $5.67$ [4.75, 6.54] mm). The condition effect was significant (Friedman $\chi^2=18.20$, $p<0.001$). Holm-corrected post hoc tests showed that AR outperformed verbal guidance and also outperformed specimen examination; the comparison between AR and specimen examination yielded $p=0.006$. \Cref{fig:end_to_end_ndi} summarizes these findings.

\subsubsection{Comparison of NDI and AR inside-out stylus measurements}
\begin{figure*}[t]
\centering
\includegraphics[width=0.98\textwidth]{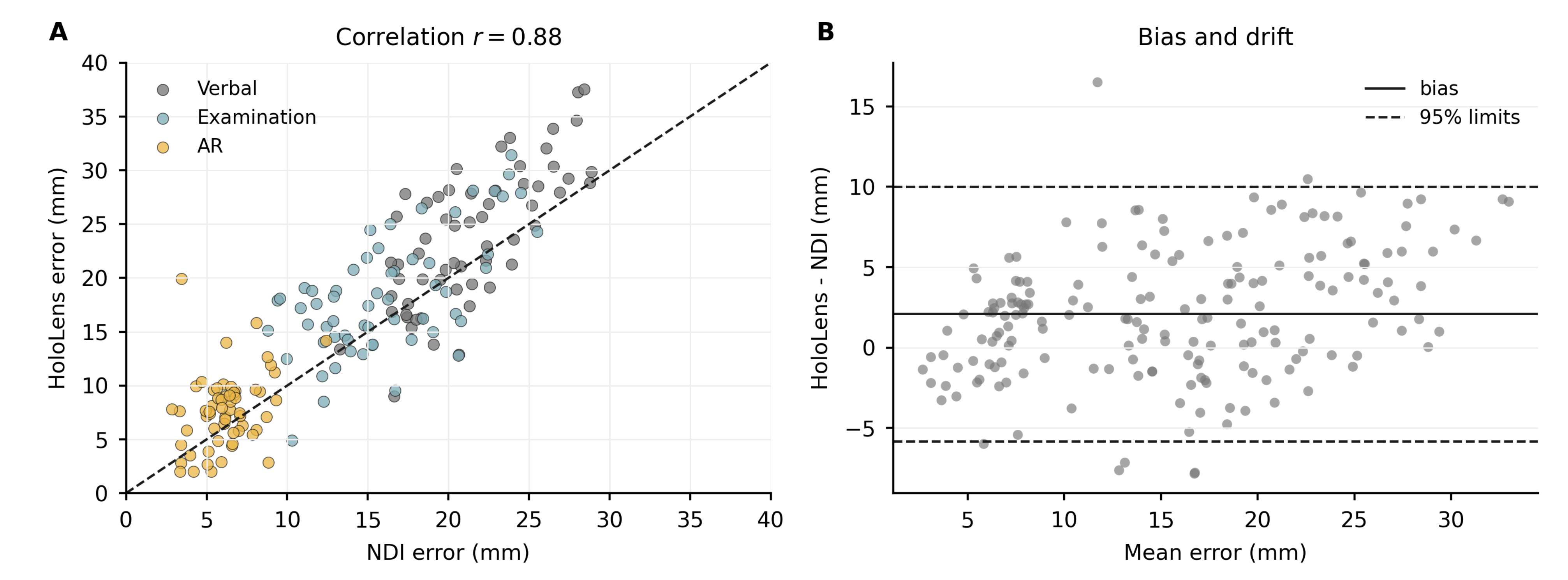}
\caption{Comparison between NDI and HoloLens HMD inside-out coordinate measurements. \textbf{(a)} Target errors measured by both coordinate sources. \textbf{(b)} HoloLens-minus-NDI bias and drift plotted against the mean error.}
\label{fig:coordinate_comparison}
\end{figure*}

We also evaluated whether AR HMD inside-out stylus tracking could support future metric collection without adding an NDI Polaris to the operating room. HoloLens HMD errors preserved the same condition ordering, with $23.92\pm6.35\mm$ for verbal guidance, $18.33\pm5.49\mm$ for specimen examination, and $7.72\pm3.44\mm$ for AR. The HoloLens-minus-NDI bias was $2.10\pm4.04\mm$, and the 95\% limits of agreement were $-5.83$ to $10.02$ mm. The overall correlation was $r=0.88$. These results indicate that HoloLens-only measurement is useful for feasibility studies, whereas NDI remains the reference for final clinical validation. \Cref{fig:coordinate_comparison} shows the scatter and agreement plots.

\input{tables/table_subjective_summary.tex}

\begin{figure*}[t]
\centering
\includegraphics[width=0.98\textwidth]{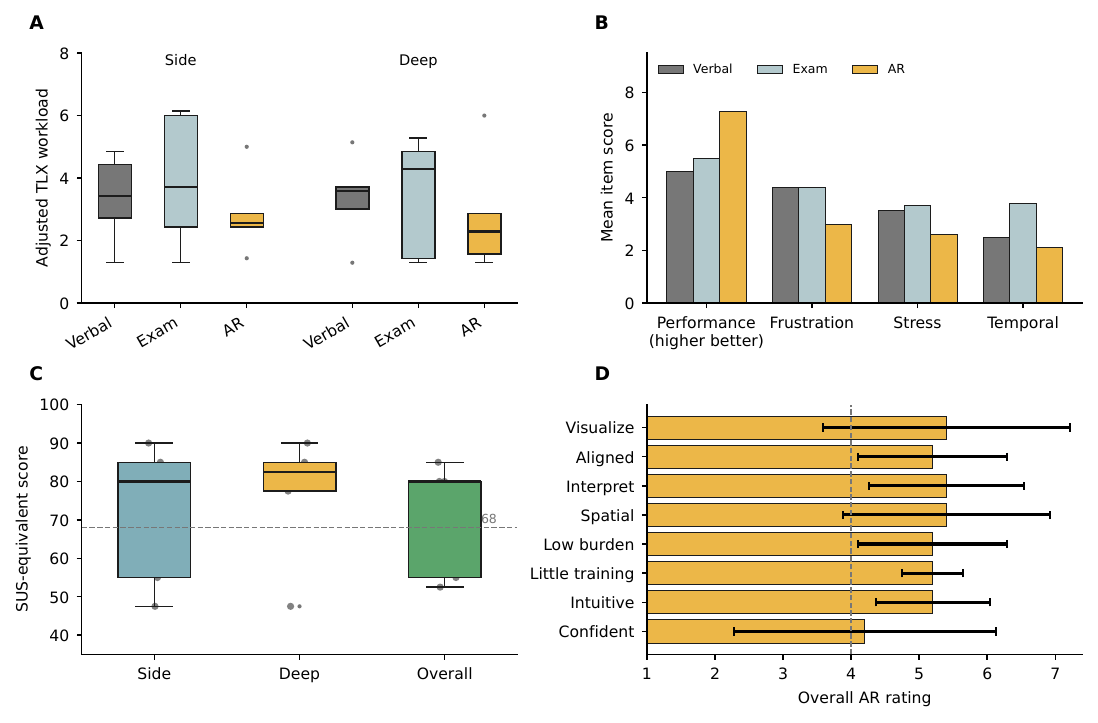}
\caption{Subjective metrics from questionnaires. \textbf{(a)} Adjusted NASA-TLX workload for peripheral and deep margins under each condition. \textbf{(b)} Mean NASA-TLX score by sub-category. \textbf{(c)} SUS-equivalent scores for side (peripheral), deep, and overall AR experience. \textbf{(d)} Summary of AR-specific questionnaire items. AR had the lowest mean workload and acceptable usability, with the strongest support for spatial understanding and self-rated performance with little training.}
\label{fig:subjective}
\end{figure*}

\subsection{Subjective measures}
Participants completed NASA-TLX workload forms \citep{hart1988nasa}, SUS usability forms \citep{reale2023medication}, and AR-specific questionnaires. We scored NASA-TLX on a 1-to-9 scale, reversing the Performance item so that higher values always indicated higher workload. For participant $i$, margin type $d\in\{\mathrm{Peripheral},\mathrm{Deep}\}$, and guidance condition $c$, the adjusted workload score was
\begin{equation}
L_{i,d,c}=\frac{1}{7}\left(M+P+T+E+F+S+10-R\right)_{i,d,c},
\end{equation}
where $M$, $P$, $T$, $E$, $F$, and $S$ are mental demand, physical demand, temporal demand, effort, frustration, and stress, and $R$ is perceived performance. We used Friedman tests to compare workload across the three guidance conditions for peripheral and deep margins \citep{friedman1937rank}. SUS was scored with the standard ten-item transformation \citep{brooke1996sus}. Let $x_{i,\ell}$ denote item $\ell$ for participant $i$. The SUS-equivalent score was
\begin{equation}
U_i=2.5\left[\sum_{\ell\in\{1,3,5,7,9\}}(x_{i,\ell}-1)+\sum_{\ell\in\{2,4,6,8,10\}}(5-x_{i,\ell})\right].
\end{equation}
Recorded zeros in the data were treated as the lowest valid anchor before scoring. AR-specific questionnaire used a 1-to-7 Likert scale and was explained to participants. We compared selected AR-specific items against the neutral value of 4 with one-sample Wilcoxon signed-rank tests \citep{wilcoxon1945individual}.

As shown in \cref{tab:subjective_summary} and \cref{fig:subjective}, the subjective results were directionally consistent with the localization findings, although the small sample limited statistical power. For peripheral margins, AR had the lowest adjusted workload mean ($2.86\,{\pm}\,1.31$), followed by verbal guidance ($3.34\,{\pm}\,1.42$), whereas specimen examination had the highest workload ($3.91\,{\pm}\,2.15$). For deep margins, AR again had the lowest mean workload ($2.80\,{\pm}\,1.89$). The Friedman tests were not significant for either peripheral or deep margins. Usability scores were acceptable for an early cadaver prototype. The SUS-equivalent score was $71.5\,{\pm}\,19.0$ after peripheral-margin AR tasks, $76.5\,{\pm}\,16.8$ after deep-margin AR tasks, and $70.5\,{\pm}\,15.5$ for the overall AR experience. Median scores were $80.0$, $82.5$, and $80.0$, respectively. Two AR-specific findings were notable. Participants rated the statement that AR supported spatial understanding on the cheek specimen at $6.0\,{\pm}\,1.0$ on a 7-point scale ($p=0.031$ against neutral, uncorrected). They also reported confidence with little training in the overall AR questionnaire at $5.2\,{\pm}\,0.4$ ($p=0.031$, uncorrected). These results suggest that the AR overlay helped participants interpret spatial relationships, even though the current system still needs refinement in user interaction design before a larger clinical study.

\subsection{Qualitative findings}
In post-study interviews, participants perceived the AR-assisted tasks as easier, of lower workload, and providing less information ambiguity than standard-of-care verbal guidance (as if from the pathologist). Participants generally found the verbal guidance (e.g. "there is a positive margin at anterior superior region near the edge.") to be imprecise and unclear.

Several participants described the tasks performed under AR guidance as smoother and more straightforward than responding to visual feedback. Some participants had difficulty seeing the edges of the visual overlay especially at off angles. Participants noted minimal lateral misalignment (1-2 mm) for the peripheral margin but more perceived misalignment (up to 1 cm) in depth for the deep margin. The AR visual display hardware was considered "finicky" whereby slight head movements could change the visibility of the displayed image. Also, some participants felt that they had to look down further than they would normally during surgery to see the AR display.

\subsection{Latency}
\label{subsec:latency_metric}
Latency was measured at the registration-update level and at the application-display-update level. For fusion attempt $a$, online fusion latency measured the time from the user triggering HMD bed fusion to acceptance of the updated transform,
\begin{equation}
L^{F}_{a}=t^{\mathrm{accept}}_{a}-t^{\mathrm{trigger}}_{a}.
\end{equation}
The fusion time was decomposed into HMD depth capture, cloud preprocessing, saliency-weighted initialization, point-to-plane refinement, and asset update. For frame $f$, the application update latency measured the delay between the latest sensor or pose timestamp and the rendered target update,
\begin{equation}
\ell^{D}_{f}=t^{\mathrm{render}}_{f}-t^{\mathrm{sensor}}_{f}.
\end{equation}
This display-update metric is not the full optical motion-to-photon latency. It captures the application-side delay that matters for updating a static surgical target. 

\begin{figure*}[t]
\centering
\includegraphics[width=0.88\textwidth]{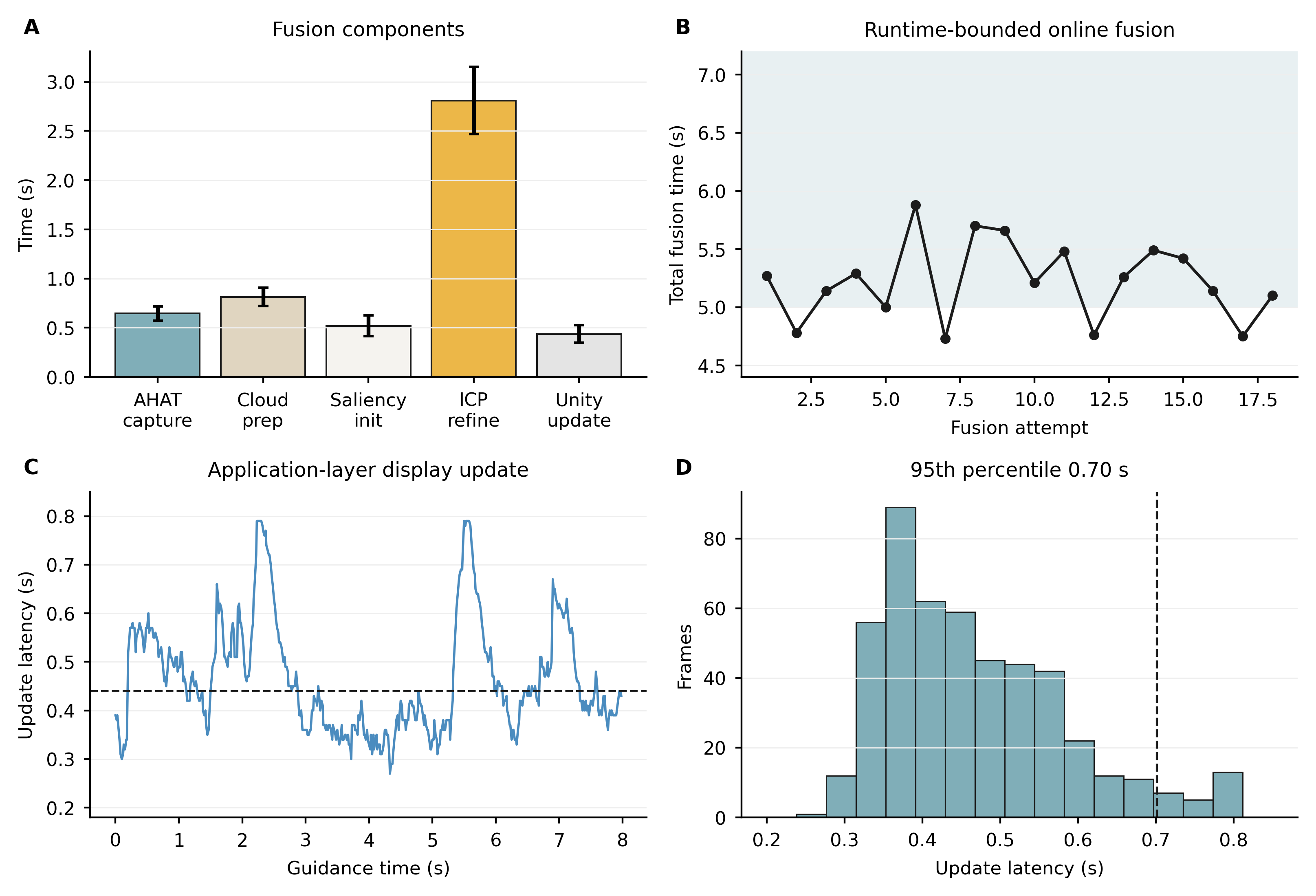}
\caption{Latency of the online AR system. \textbf{(a)} Fusion-runtime decomposition. \textbf{(b)} Online fusion times across repeated attempts, with the shaded band marking the 5 to 8 s target. \textbf{(c)} Latency trace during guidance. \textbf{(d)} Display-update latency distribution.}
\label{fig:latency}
\end{figure*}

Across two cadavers, the total online fusion time was $5.23\,{\pm}\,0.34\s$, with a median of $5.23$~s and a 95th percentile of $5.73$~s. The frame-level application update latency had a mean of $0.47\,{\pm}\,0.11\s$, a median of $0.44$~s, and a 95th percentile of $0.70$~s. \Cref{fig:latency} summarizes these timing results.

\section{Discussion}
\label{sec:discussion}

We developed a marker-free system and workflow for positive margin localization during tumor re-resection. The primary finding was that AR guidance decreased end-to-end localization error by over 70\%, from $21.40\pm3.84\mm$ with verbal guidance and $16.09\pm4.30\mm$ with verbal plus specimen examination to $6.19\pm1.79\mm$ with AR guidance. This improvement supports the clinical motivation of the system: instead of asking the surgeon to mentally transfer a positive margin from a detached, deformed specimen back to the patient (a task that is especially difficult when feedback is limited to a verbal anatomic description), the proposed AR-enhanced workflow returns the specimen-derived margin as a spatial target on the resection bed.

This advantage is important because the current clinical workflow can require surgeons to compensate for uncertainty by taking a larger re-resection, which may remove healthy tissue, or by sampling an imprecise site, which may miss residual disease. AR guidance may therefore improve the success rate of the re-resection, reduce unnecessary tissue removal, shorten time spent re-orienting the specimen and resection bed, support better functional and cosmetic recovery, and reduce the risk of future surgery. These benefits are especially relevant in head and neck surgery, where a few millimeters can affect speech, swallowing, facial appearance, or reconstruction options.

An important interpretation of the end-to-end metric is that it is not simply the sum of the component errors. In an optical see-through AR system, the user sees the green predicted margin target while also seeing the true resection bed through the visor. The user naturally places the tool tip onto the physical resection bed surface rather than blindly selecting a virtual point in space. Therefore, the end-to-end error measures human-system performance, focusing on the user's ability to interpret the AR cue and locate the intended patient-side margin. This is the clinically important task. At the same time, component-level metrics remain necessary because they show where the system can improve. Deformation accuracy, deformed specimen-to-bed registration, marker-free fusion, latency, and subjective usability each explain a different part of the pipeline. This evaluation design may also be useful for other computer-assisted intervention systems, especially AR systems where final performance depends on both registration accuracy and user interpretation.

The component results show that marker-free fusion is accurate, but deformation and heavy occlusion remain important sources of error. No significant paired difference in fusion error was detected between the marker-free and marker-based methods on either cadaver. The combined
marker-free error was $2.15\,{\pm}\,0.87\mm$, with a maximum of $4.41\mm$. This is encouraging because marker-based systems add setup burden, line-of-sight constraints, and sterile-field complexity. However, heavy draping increased the marker-free fusion median error to $3.14\mm$, showing that limited visible anatomy can reduce fusion reliability. Deformation also varied by specimen, with TRE of $7.63\,{\pm}\,3.74\mm$ for the cheek and $3.72\,{\pm}\,1.02\mm$ for the scalp. Future improvements should therefore focus on stronger deformation models, better contour extraction, more robust resection bed segmentation, and confidence-based fusion checks under draped conditions.

The workflow was designed to be clinically integrated rather than fully separate from standard margin management. The surgeon still performs the initial resection, the specimen still enters the pathology workflow, and the pathologist still labels the margin. The main added manual steps are bead-suture placement and 3D specimen scanning, while deformation, registration, and AR preparation are intended to run during the pathology waiting period. In the current prototype, these added steps took approximately 10 minutes. The 3D scanning step remains a practical limitation and a future research direction. Faster scanning, automated mesh reconstruction, improved specimen holders, and direct integration with pathology annotation software could reduce the added time and make the workflow easier to adopt.

This study has several limitations. The evaluation used two cadaveric sites and five participants, so the results demonstrate feasibility rather than clinical efficacy. Several steps also remain manual or semi-manual, including segmentation, fiducial selection, and label placement. Future work will address identified or other user interaction improvements, including opacity control, target visualization modalities, and registration-confidence warnings, to prepare for a larger clinical trial. A major future direction is tongue and oral cavity re-resection, where tissue preservation is especially important and where more complex deformation will make accurate AR guidance both more challenging and potentially more valuable.

\section{Conclusion}
\label{sec:conclusion}

We presented a marker-free AR system for positive margin localization during tumor re-resection surgery. The system combines contour-constrained deformable registration, residual specimen-to-bed registration, drape-aware marker-free bed-to-AR fusion, and bed-anchored rendering of positive margin targets. In a cadaveric cheek and scalp specimen study, AR guidance significantly reduced localization error compared with verbal guidance and specimen examination, while no significant difference in fusion error was detected between the marker-free and marker-based methods.

The end-to-end result should be interpreted as user performance in the AR-guided clinical task, not as a simple accumulation of component errors. The surgeon sees both the virtual target and the real resection bed, and uses the overlay to decide where to place the tool on physical tissue. These results support the feasibility of marker-free AR-guided re-resection, while also identifying deformation robustness, drape occlusion, scanning time, and user interaction design as key targets before larger clinical validation. Thus, this work moves us toward a future in which surgeons can remove cancer more completely, preserve more healthy tissue, and further improve lives for patients around the world.

\section{Funding}
This work was supported by the National Institute of Biomedical Imaging and Bioengineering (NIBIB) of the NIH under grant R01EB037685.  This work was also supported in part by the NIBIB-NIH grant T32EB021937.

\section{Declaration of Interest}
The authors declare no conflict of interest.

% Uncomment and use as the case may be
%\begin{theorem} 
%\end{theorem}

% Uncomment and use as the case may be
%\begin{lemma} 
%\end{lemma}


\printcredits\textit{}

\bibliographystyle{cas-model2-names}
\bibliography{marker_free_ar_revised_refs}



\end{document}

%% file: tables/table_setup.tex
\begin{table*}[t]
\centering
\caption{Experimental setup and participant demographics.}
\label{tab:setup}
\small
\renewcommand{\arraystretch}{1.12}
\setlength{\tabcolsep}{5pt}

\begin{tabularx}{\textwidth}{@{}
>{\hsize=0.70\hsize\linewidth=\hsize\raggedright\arraybackslash}X
>{\hsize=1.10\hsize\linewidth=\hsize\raggedright\arraybackslash}X
>{\hsize=1.20\hsize\linewidth=\hsize\raggedright\arraybackslash}X
@{}}
\toprule
\rowcolor{gray!18}
\textbf{Category} & \textbf{Value} & \textbf{Notes} \\
\midrule
\rowcolor{gray!7}
Participants & $n=5$, age 33 $\pm$ 5 years & 3 male and 2 female. All reported some prior familiarity with VR and AR. \\
Clinical background & 3 residents, 1 fellow, and 1 attending surgeon & 4 participants were in ENT and 1 participant was in general surgery on a non-ENT rotation. \\
\rowcolor{gray!7}
Cadaver sites & Specimen 1 cheek; Specimen 2 scalp & Each specimen contained 4 peripheral-margin targets and 2 deep-margin targets. \\
Guidance conditions & Verbal, verbal with specimen examination, and AR guidance & Task order followed the cadaver evaluation protocol with NASA-TLX after tasks and SUS/AR questionnaires after AR blocks. \\
\bottomrule
\end{tabularx}
\end{table*}

%% file: tables/table_subjective_summary.tex
\begin{table*}[t]
\centering
\small
\caption{Subjective workload, usability, and AR-specific ratings. NASA-TLX values use a 1 to 9 adjusted workload scale after reversing the performance item. SUS-equivalent values use the standard 0 to 100 scoring. AR-specific items use a 1 to 7 Likert scale, and one-sample Wilcoxon tests compare against the neutral score of 4. Notable AR results are bolded. }
\label{tab:subjective_summary}
\renewcommand{\arraystretch}{1.12}
\setlength{\tabcolsep}{4pt}

\begin{tabularx}{\textwidth}{@{}
>{\hsize=1.03\hsize\linewidth=\hsize\raggedright\arraybackslash}X
>{\hsize=1.41\hsize\linewidth=\hsize\raggedright\arraybackslash}X
>{\hsize=0.71\hsize\linewidth=\hsize\raggedright\arraybackslash}X
>{\hsize=0.98\hsize\linewidth=\hsize\raggedright\arraybackslash}X
>{\hsize=0.87\hsize\linewidth=\hsize\raggedright\arraybackslash}X
@{}}
\toprule
\rowcolor{gray!18}
\textbf{Metric} & \textbf{Group} & \textbf{Mean $\pm$ SD} & \textbf{Median [IQR]} & \textbf{Test} \\
\midrule
\rowcolor{gray!7}
\makecell[l]{NASA-TLX adjusted\\workload} & Peripheral Verbal & 3.34 $\pm$ 1.42 & 3.43 [2.71, 4.43] & \makecell[l]{Friedman\\$p=0.623$} \\
\makecell[l]{NASA-TLX adjusted\\workload} & Peripheral Examination & 3.91 $\pm$ 2.15 & 3.71 [2.43, 6.00] & \makecell[l]{Friedman\\$p=0.623$} \\
\rowcolor{gray!7}
\makecell[l]{NASA-TLX adjusted\\workload} & Peripheral AR & \textbf{2.86 $\pm$ 1.31} & \textbf{2.57 [2.43, 2.86]} & \makecell[l]{Friedman\\$p=0.623$} \\
\makecell[l]{NASA-TLX adjusted\\workload} & Deep Verbal & 3.34 $\pm$ 1.39 & 3.57 [3.00, 3.71] & \makecell[l]{Friedman\\$p=0.949$} \\
\rowcolor{gray!7}
\makecell[l]{NASA-TLX adjusted\\workload} & Deep Examination & 3.43 $\pm$ 1.92 & 4.29 [1.43, 4.86] & \makecell[l]{Friedman\\$p=0.949$} \\
\makecell[l]{NASA-TLX adjusted\\workload} & Deep AR & \textbf{2.80 $\pm$ 1.89} & \textbf{2.29 [1.57, 2.86]} & \makecell[l]{Friedman\\$p=0.949$} \\
\rowcolor{gray!7}
SUS equivalent & Peripheral & 71.5 $\pm$ 19.0 & 80.0 [55.0, 85.0] & \makecell[l]{vs.\ 68 benchmark\\$p=0.406$} \\
SUS equivalent & Deep & 76.5 $\pm$ 16.8 & 82.5 [77.5, 85.0] & \makecell[l]{vs.\ 68 benchmark\\$p=0.219$} \\
\rowcolor{gray!7}
SUS equivalent & Overall & \textbf{70.5 $\pm$ 15.5} & \textbf{80.0 [55.0, 80.0]} & \makecell[l]{vs.\ 68 benchmark\\$p=0.500$} \\
AR-specific item & \makecell[l]{Cheek specimen, supported\\spatial understanding} & 6.0 $\pm$ 1.0 & 6.0 [5.0, 7.0] & \makecell[l]{one-sample\\$p=0.031$} \\
\rowcolor{gray!7}
AR-specific item & \makecell[l]{Overall, confident\\with little training} & 5.2 $\pm$ 0.4 & 5.0 [5.0, 5.0] & \makecell[l]{one-sample\\$p=0.031$} \\
AR-specific item & Overall, easy to interpret & 5.4 $\pm$ 1.1 & 5.0 [5.0, 6.0] & \makecell[l]{one-sample\\$p=0.062$} \\
\bottomrule
\end{tabularx}
\end{table*}